\documentclass[letterpaper, 10 pt, conference]{ieeeconf}
\IEEEoverridecommandlockouts                            
\title{Energy-Based Dynamical Models for \\ Neurocomputation, Learning, and Optimization}

\author{Arthur N. Montanari, Francesco Bullo, Dmitry Krotov, and Adilson E. Motter
\thanks{This work was partially supported by the ARO Grant No.\ W911NF-24-1-0228.}
\thanks{
ANM is with the Center for Network Dynamics and the Department of Physics \& Astronomy, Northwestern University, Evanston, IL 60208, USA.
FB is with the Center for Control, Dynamical Systems, and Computation and the Department of Mechanical Engineering, University of California at Santa Barbara, Santa Barbara, CA 93106, USA.
DK is an Independent Researcher, Cambridge, MA 02141, USA.
AEM is with the Center for Network Dynamics, the Department of Physics \& Astronomy, the Northwestern Institute on Complex Systems, and the Department of Engineering Sciences \& Applied Mathematics, Northwestern University, Evanston, IL 60208, USA. 
Corresponding author: arthur.montanari@northwestern.edu.
}
}

\usepackage[noadjust]{cite} 		  
\usepackage{xcolor}       
\usepackage{hyperref}	  
\hypersetup{colorlinks=true,linkcolor=blue,citecolor=blue,breaklinks={true}} 
\usepackage{comment}

\usepackage{graphicx}
\DeclareGraphicsExtensions{.eps,.png,.jpg,.jpeg,.bmp,.gif,.pdf}

\usepackage{tabularx,booktabs}	
\usepackage{multirow}	

\usepackage{amsmath}      		
\usepackage{amssymb}	  		
\usepackage{bm}		      		
\usepackage{physics}	  		
\usepackage{soul}				

\newcommand{\R}{\mathbb{R}}		
\newcommand{\transp}{\mathsf{T}}					

\DeclareMathOperator*{\argmin}{arg\,min}

\usepackage{amsthm}
\newtheorem{defin}{Definition}
\newtheorem{thm}{Theorem}
\newtheorem{cor}{Corollary}

\theoremstyle{definition}

\newtheorem{rem}{Remark}

\newtheorem{result}{Result}

\renewcommand{\QED}{\QEDopen}

\begin{document}
\maketitle
\thispagestyle{empty}
\pagestyle{empty}

\begin{abstract}      
Recent advances at the intersection of control theory, neuroscience, and machine learning have revealed novel mechanisms by which dynamical systems perform computation. These advances encompass a wide range of conceptual, mathematical, and computational ideas, with applications for model learning and training, memory retrieval, data-driven control, and optimization. This tutorial focuses on neuro-inspired approaches to computation that aim to improve scalability, robustness, and energy efficiency across such tasks, bridging the gap between artificial and biological systems. Particular emphasis is placed on \textit{energy-based dynamical models} that encode information through gradient flows and energy landscapes. We begin by reviewing classical formulations, such as continuous-time Hopfield networks and Boltzmann machines, and then extend the framework to modern developments. These include dense associative memory models for high-capacity storage, oscillator-based networks for large-scale optimization, and proximal-descent dynamics for composite and constrained reconstruction. The tutorial demonstrates how control-theoretic principles can guide the design of next-generation neurocomputing systems, steering the discussion beyond conventional feedforward and backpropagation-based approaches to artificial intelligence.
\end{abstract}

\section{Introduction}

\begin{figure}[!b]
    \centering
    \includegraphics[width=\linewidth]{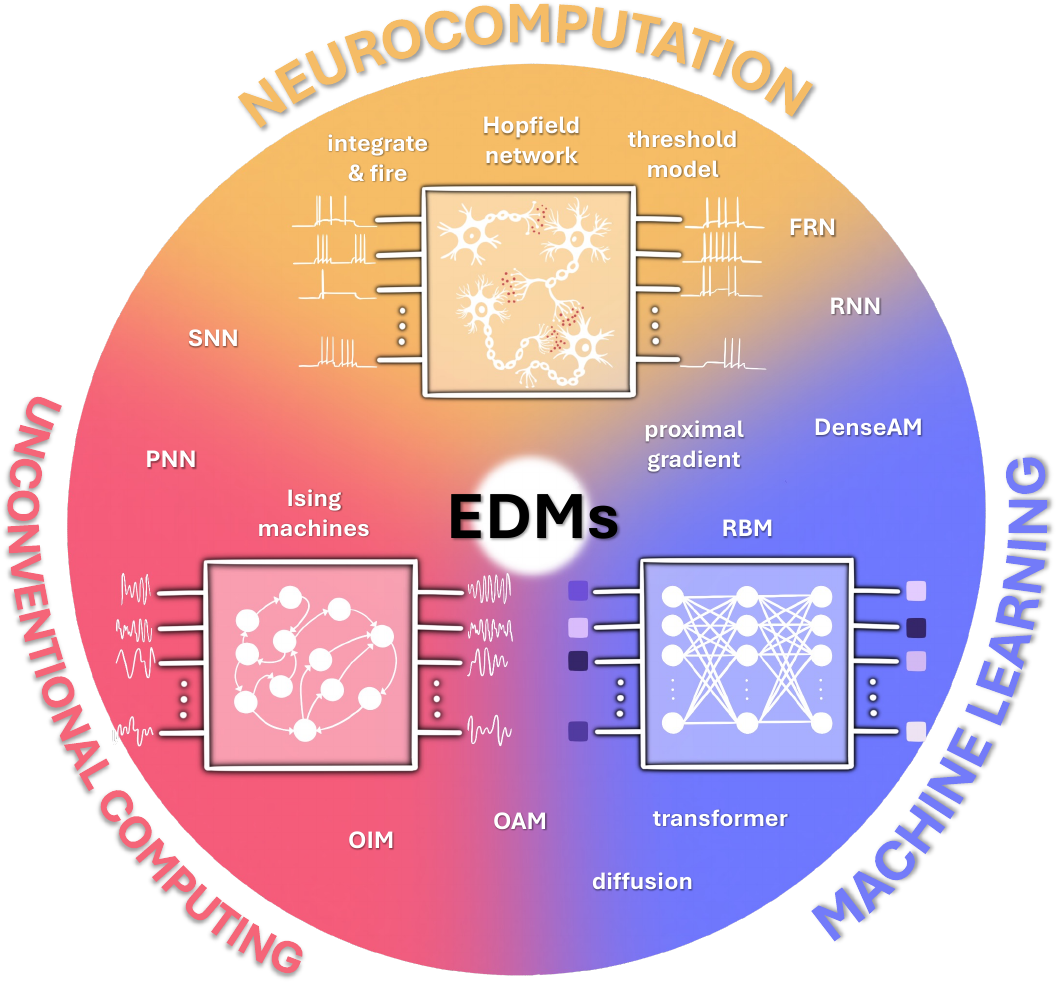}
    \caption{Dynamical systems used in neurocomputation, machine learning, and unconventional computing that admit EDM formulations (the acronyms are defined throughout the text). In biological neural systems, computation is most commonly interpreted as emerging from spiking activity and complex recurrent circuitry. In contrast, machine learning models simplify this representation to layered neural architectures with feedforward signal propagation and static nonlinear activations, which facilitate training and inference. Unconventional computing, on the other hand, exploits the intrinsic physical dynamics of analog hardware to perform computation; many such systems, including electronic oscillators, photonic devices, resistor networks, and spintronics, exhibit oscillatory or nonlinear dynamical behavior. The EDM formalism presented in this tutorial provides a unifying framework to study and design neurocomputational models across these three domains.}
    \label{fig.diagram}
\end{figure}

Deep learning has transformed artificial intelligence (AI), leading to breakthroughs in computer vision, natural language processing, and scientific discovery \cite{lecun2015deep,bengio2017deep}. Most of these models\textemdash ranging from feedforward neural networks (FNNs) to the now widely used transformers and diffusion models\textemdash are trained via backpropagation and gradient-based optimization methods that have been highly optimized for massive datasets, deep layered architectures, and parallel hardware. Despite their remarkable empirical success, deep-learning systems exhibit fundamental limitations in \textit{interpretability}, \textit{reliability}, \textit{scalability}, and \textit{biological plausibility}. The internal representations learned by deep neural networks are difficult to analyze, as the relationship between learned features, model parameters, and the resulting computation lacks a mechanistic interpretation \cite{lipton2016mythos}. The black-box nature of these models complicates their output regulation and limits the reliability of their predictions in safety-critical applications \cite{amodei2016concrete,gyevnar2025ai}. Moreover, training and inference require enormous computational resources, resulting in high energy demand and poor scalability \cite{desislavov2023trends}. These costs derive from backpropagation, which requires repeated forward and backward passes through the network and synchronized updates across a large number of parameters. The resulting learning mechanisms thus bear little resemblance to biological neural computation, in which learning is believed to arise from local, continuous, and recurrent neural interactions \cite{lillicrap2020backpropagation}. These limitations motivate the search for alternative computational frameworks.

Energy-based dynamical models (EDMs) provide a unifying framework for neurocomputation, machine learning, and unconventional computing (Fig.~\ref{fig.diagram}). In EDMs, computation is performed through the evolution of a dynamical system whose trajectories $\bm x(t)\in\R^N$ monotonically decrease a scalar function $\mathcal E(\bm x) : \R^N \mapsto \R$, commonly referred to as an \textit{energy}, \textit{Lyapunov}, or \textit{objective} function. A canonical example is the gradient-descent system $\dot{\bm x} = -\nabla \mathcal E(\bm x)$.  
This representation provides a mechanistic interpretation of computation, where information is encoded in the properties of the energy landscape. 
Learning corresponds to the process of sculpting the landscape so that its local minima correspond to desired solutions of inference, memory retrieval, or optimization tasks. Computation then emerges as the system autonomously relaxes toward equilibria, limit cycles, or other attractors corresponding to those solutions. 
%
EDMs are thus highly amenable to control-theoretic tools, since their stability, convergence, and robustness can be characterized and designed through the energy function itself, improving reliability. This perspective provides a principled framework for addressing several of the limitations discussed above.

EDMs also connect several historically distinct research domains: 
\begin{enumerate}
    \item In \textit{theoretical neuroscience}, mathematical models of brain function have long described computation as a process in which neural activity evolves to minimize an internal energy, objective, or prediction error. Early examples include Hopfield's associative memory networks \cite{hopfield1982neural,hopfield1984neurons,gerstner2014neuronal}, competitive learning \cite{grossberg1976adaptive,cohen1983absolute}, and predictive coding \cite{rao1999predictive,friston2018does}. In this context, Karl Friston's free-energy principle \cite{KF:10} proposes that biological systems maintain an internal model of the environment and continuously adjust their neural activity to reduce the mismatch (also known as ``surprise'') between predicted and observed sensory inputs. 
    
    \item In \textit{machine learning}, generative models seek to learn the probability distribution underlying observed data. Boltzmann machines \cite{hinton1983optimal,ackley1985learning} and other energy-based models \cite{lecun2006tutorial,grathwohl2019your,arbel2020generalized} represent these probability distributions through an energy function defined over the system variables (e.g., $\mathbb P(\bm x) \propto \exp{-\mathcal E(\bm x)}$). In this formulation, data-like states have low energy and are thus more likely to be sampled by the trained model. 

    \item In \textit{unconventional computing}, physical systems are engineered so that their dynamical evolution performs computation. An optimization problem is encoded directly into the interactions of a physical device, whose energy (i.e., Hamiltonian) matches the objective function of interest \cite{du2025physical}. The device then relaxes according to its intrinsic dynamics, driving the physical state toward low-energy configurations that represent candidate solutions of the optimization problem. Examples include Ising machines \cite{inagaki2016coherent,goto2019combinatorial,mohseni2022ising}, oscillator networks \cite{wang2019oim,mallick2020using,wang2024training}, trainable physical neural networks (PNNs) \cite{stern2021supervised,stern2025physical}, and other analog computing hardware.
\end{enumerate}

This tutorial introduces the mathematical foundations of EDMs, connecting computation, stability analysis, and control theory. As canonical examples, we first review the continuous-time Hopfield network as a deterministic EDM for associative memory and the continuous-time Boltzmann machine as a stochastic EDM for generative modeling (Sec.~\ref{sec.background}). We then discuss learning mechanisms for EDMs that are biologically plausible, local, cooperative, and unsupervised, overcoming the main limitations of the backpropagation algorithm (Sec.~\ref{sec.learning}).
Building on these foundations, we introduce modern EDM approaches aimed at improving storage capacity, enabling efficient hardware implementation, and scaling optimization for large problems. These include dense associative memory models \cite{krotov2016dense,krotov2018dense}, a generalization of Hopfield networks for high-capacity storage (Sec.~\ref{sec.dam}); oscillator-based network models \cite{nishikawa2004capacity,wang2019oim,allibhoy2025global} for error-free memory retrieval and large-scale combinatorial optimization (Sec.~\ref{sec.oscillator}); and proximal gradient descent neural networks \cite{SHM-MRJ:21,AG-AD-FB:24d} for sparse and constrained optimization (Sec.~\ref{sec.proximal}).
Together, this tutorial provides a cohesive perspective on how dynamical principles can be exploited for the design of computational models that are robust, interpretable, and scalable.

\section{Energy-Based Dynamical Models}
\label{sec.background}

An EDM takes the form
\begin{equation}
    \dot{\bm x} = \bm f(\bm x; \bm\theta),
    \label{eq.generaledm}
\end{equation}
where $\bm x\in\R^N$ denotes the system state (e.g., neural activities, phases, and firing rates) and $\bm\theta$ denotes parameters (e.g., synaptic weights). The defining property of an EDM is the existence of a scalar energy function $\mathcal E : \R^N\mapsto\R$ satisfying
\begin{equation}
    \dot{\mathcal E}(\bm x) = \nabla \mathcal E(\bm x)^\transp \dot{\bm x} \leq 0.
    \label{eq.energyderiv}
\end{equation}
When this condition holds, $\mathcal E$ serves as a Lyapunov function candidate, and the system is dissipative with respect to $\mathcal E$.
A canonical example is the gradient-flow system
\begin{equation}
    \dot{\bm x} = -\nabla \mathcal E(\bm x;\bm \theta),
    \label{eq.edm}
\end{equation}
for which $\dot {\mathcal E}(\bm x)=-\|\nabla \mathcal E(\bm x)\|^2\le 0$. More generally, EDMs can be represented by:
\begin{enumerate}
    \item \textit{preconditioned} gradient flows,
    \begin{equation}
        \dot{\bm x} = -M(\bm x)\nabla \mathcal E(\bm x),
        \label{eq.precond}
    \end{equation}
    where $M(\bm x)=M(\bm x)^\transp \succeq 0$ is a state-dependent matrix;

    \item \textit{stochastic} gradient flow, such as the overdamped Langevin dynamics
    \begin{equation}
        {\rm d}\bm x_t = -\nabla \mathcal E(\bm x_t) {\rm d}t + \sqrt{2T}{\rm d}\bm w_t,
        \label{eq.stogradflow}
    \end{equation}
    where $\bm w_t$ is a standard Wiener process and $T>0$ controls the noise intensity, enabling sampling-based inference and generative modeling;
    
    \item \textit{projected} 
    gradient flow arising in constrained and composite optimization,
    \begin{equation}
        \dot{\bm x} = \Pi_{\mathcal C}\!\left(-\nabla \mathcal E(\bm x)\right),
        \label{eq.projgradflow}
    \end{equation}
    where 
    $\Pi_{\mathcal C}$ denotes the projection onto a constraint set $\mathcal C$.
\end{enumerate}

Many computational problems in machine learning can be formulated as an optimization problem of the form
\begin{equation}
    \min_{\bm x} \mathcal L(\bm x;\bm\theta),
    \label{eq.optimization}
\end{equation}
where $\mathcal L$ encodes a task-specific loss function. EDMs can naturally act as solvers to such problems, where the function $\mathcal E$ plays a dual role relating \textit{optimization} and \textit{stability}. On the one hand, $\mathcal E$ encodes the computational objective, in the sense that its minima correspond to solutions of the task (e.g., $\mathcal E=\mathcal L$ in simple settings). On the other hand, $\mathcal E$ serves as a Lyapunov function guaranteeing stability and convergence of the dynamical system. Unlike standard pipelines in machine learning, where inference is an explicit input-output map (e.g., FNNs), EDMs implement \textit{implicit computation}: Solutions to the optimization problem \eqref{eq.optimization} are obtained as the asymptotic behavior of a dynamical system. For the gradient flow \eqref{eq.edm} with $\mathcal E=\mathcal L$, equilibria $\bm x^\star$ satisfy $\nabla \mathcal L(\bm x^\star;\bm\theta)=0$, and hence strict local minima of $\mathcal L$ (i.e., $\nabla^2 \mathcal L(\bm x^\star)\succ 0$) correspond to asymptotically stable equilibria. Thus, there is a direct connection between the dynamics of the EDM \eqref{eq.edm} and the solutions of the optimization problem \eqref{eq.optimization}.

A central feature of EDMs is the separation of inference and learning across different timescales. \textit{Inference} or \textit{retrieval} corresponds to the fast evolution of the state $\bm x$ under fixed parameters $\bm \theta$, during which the EDM~\eqref{eq.edm} relaxes from an initial condition $\bm x(0)$ towards an equilibrium $\bm x^*$ (or, more generally, an attractor) encoding a solution to the optimization problem. \textit{Learning} occurs on a slower time scale and governs the adaptation of parameters, 
\begin{equation}
    \dot{\bm\theta} = \bm g(\bm\theta;\bm x),
\label{eq.paramlearning}
\end{equation}
which parameterizes the energy landscape itself. In neurocomputation, the update rule $\bm g$ is often local (depending only on pre- and post-synaptic variables) and can be interpreted as reshaping $\mathcal E(\bm x;\bm\theta)$ so that desired states become stable equilibria (memory storage) or so that the induced stationary distribution matches the observed data (generative modeling). In what follows, we make this connection explicit for Hopfield networks and Boltzmann machines, respectively. These examples focus on the inference stage, while the learning mechanisms are discussed in Sec.~\ref{sec.learning}.

\subsection{Continuous-Time Hopfield Networks}
\label{sec.hopfield}

A classic example of EDMs for neurocomputation is the Hopfield neural network, originally introduced as a model of \textit{associative memory} in the brain \cite{hopfield1982neural}.
Let $W\in\R^{N\times N}$ be a synaptic weight matrix, where $N$ is the number of neurons in the network.
The continuous-time dynamics of the Hopfield model \cite{hopfield1984neurons} are described by
\begin{equation}
    \tau\dot{\bm x} = -D\bm x + W \bm \Phi(\bm x) + B\bm u,
    \label{eq.cthopfield}
\end{equation}
where $\tau>0$ is the time constant, $D = \operatorname{diag}(d_1,\ldots,d_N)$ is a diagonal matrix of dissipation rates $d_i>0$, $\bm u\in\R^M$ represents a (possibly time-varying) input or bias, $B\in\R^{N\times M}$ is the input matrix, and $\bm \Phi:\R^N\mapsto\R^N$ is an activation function applied elementwise, with $\bm \Phi(\bm x)=(\Phi(x_1),\dots,\Phi(x_N))^\top$. By assumption, $\Phi : \R\mapsto\R$ satisfies the  properties:
\begin{enumerate}
    \item[i)] weakly increasing and non-expansive;
    \item[ii)] sign preserving (i.e., $\Phi(x)>0$ for $x>0$ and $\Phi(x)<0$ for $x<0$);
    \item[iii)] unbounded integral (i.e., $\lim_{z\rightarrow\infty}\int_0^z \Phi(x){\rm d}x = \infty$).
\end{enumerate}
Examples include saturation functions, such as the hyperbolic tangent, the sigmoidal function, and ReLU. Importantly, $W$ can encode arbitrary networks, including those with cycles (feedback loops). Therefore, in contrast to layered FNNs (cf. Eq.~\eqref{eq.feedforward}), the Hopfield model constitutes a \textit{recurrent} neural network (RNN).

\begin{rem}
RNNs are widely used in AI \cite{yu2019review,krotov2016dense,ramsauer2020hopfield}, including language modeling, speech recognition, and time-series prediction. 
These models maintain an evolving state that acts as a memory of past inputs, capturing temporal dependencies absent in purely FNN architectures. Yet, training RNNs can be challenging due to well-known issues such as exploding/vanishing gradients \cite{ribeiro2020beyond}. Several architectures have been proposed to mitigate these problems, including long short-term memory networks (LSTMs) \cite{hochreiter1997long}, gated recurrent units (GRUs) \cite{cho2014learning}, and neural ODE models \cite{chen2018neural}.
\end{rem}

\begin{figure*}[t]
    \centering
    \includegraphics[width=0.92\linewidth]{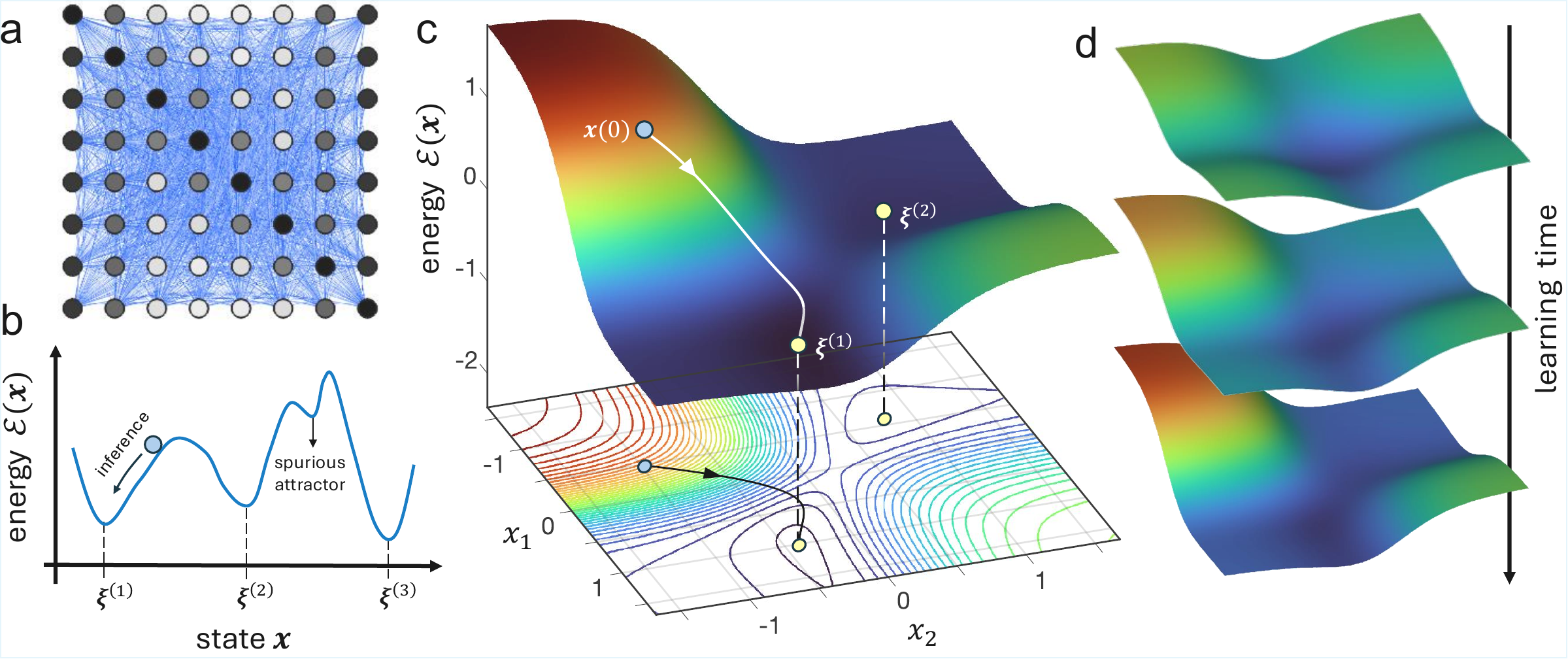}
    \caption{Continuous-time Hopfield neural network for associative memory. 
    (a) Neural network, where nodes represent individual neurons $x_i$ and edges represent pairwise interactions $W_{ij}$. The activity of each neuron encodes the pixel intensity of a grayscale image, forming a memory pattern $\bm\xi^{(\mu)}$ corresponding to the letter N. 
    (b) Schematic of the energy landscape of a high-dimensional Hopfield network. Stored memory patterns $\bm\xi^{(\mu)}$ are encoded as local minima of $\mathcal E(\bm x)$, and inference occurs as the network state $\bm x(t)$ relaxes from an initial condition $\bm x(0)$ toward a nearby minimum through gradient descent. The energy landscape may contain undesired local minima, known as spurious attractors.
    (c) Energy landscape $\mathcal E(\bm x)$ of a Hopfield network with $N=2$ neurons and $K = 2$ patterns. The contour levels are shown in the $(x_1,x_2)$ plane. 
    (d) Hebbian learning adjusts the synaptic weights $W_{ij}$ so that the energy landscape can encode (store) the desired memory patterns as local minima. In this example, the energy landscape is initialized with random weights, yielding a monostable system. As the weights evolve according to Oja's rule, the energy function encodes the desired patterns.}
    \label{fig.hopfield}
\end{figure*}

A common interpretation in neuroscience is that $x_i$ represents the membrane potentials and $\Phi(x_i)$ its firing rate. Each stable equilibrium $\bm x^*$ corresponds to a \textit{memory pattern} $\bm\xi^{(\mu)}\in\R^N$, indexed by $\mu=1,\ldots,K$, that the network is designed to \textit{store} and later \textit{retrieve}. Memory patterns are thus encoded as vectors; when each component $\xi^{(\mu)}_i$ is associated with a pixel, a frequency coefficient, or a token feature, the resulting pattern $\bm\xi^{(\mu)}$ can represent, respectively, an image, a sound, or a symbolic object like a word.
Learning consists of tuning the weights $W_{ij}$ so that prescribed patterns $\bm\xi^{(\mu)}$ become stable equilibria $\bm x^*$, while inference (or memory retrieval) corresponds to relaxing system \eqref{eq.cthopfield} for a certain set of inputs $\bm u$, or initial conditions $\bm x(0)$, toward the desired memory attractor. In practice, a Hopfield model is designed to be multistable, allowing $K\geq 2$ patterns to be stored simultaneously.
Fig.~\ref{fig.hopfield} illustrates the Hopfield network model.

Under certain conditions, the Hopfield network can be described as an EDM with the energy function \cite{hopfield1984neurons,bullo2025lecturesneural}:
\begin{equation}
\begin{aligned}
    \mathcal E(\bm x) =& -\frac{1}{2} \bm\Phi(\bm x)^\transp W \bm\Phi(\bm x) 
       + (D\bm x - B\bm u)^\transp \bm\Phi(\bm x) \\
       &- \sum_{i=1}^N d_i \int_{0}^{x_i} \Phi (w) {\rm d}w.
    \label{eq.hopfield_energy}
\end{aligned}
\end{equation}

\begin{thm} \label{thm.hopfieldlyapunov}
If $W=W^\top$ (symmetric weights) and $\Phi$ is continuously differentiable, then
\begin{equation}
    \dot{\mathcal E}(\bm x(t)) = -\frac{1}{\tau} \sum_{i=1}^N d_i \Phi'(x_i(t)) \big(\dot x_i(t)\big)^2 \leq 0, \,\, \forall \bm x,
    \label{eq.hopfielddecrease}
\end{equation}
where $\Phi'$ denotes the derivative of $\Phi$.
\end{thm}

\noindent
Thus, $\mathcal E$ is nonincreasing along trajectories of system \eqref{eq.cthopfield}. By LaSalle's invariance principle, every trajectory approaches $\{\bm x: \dot{\mathcal E}(\bm x)=0\}$. If $\Phi'(x_i)>0$, $\forall x_i$, the set of critical points of $\mathcal E$ coincides precisely with the set of equilibria of \eqref{eq.cthopfield}. The Hopfield model can thus be written as the preconditioned gradient flow \eqref{eq.precond}, where $M(\bm x) = \frac{1}{\tau}\operatorname{diag}(1/\bm\Phi'(\bm x))$. This EDM representation provides a direct computational interpretation: memory retrieval corresponds to gradient descent on an energy landscape, which is directly ``shaped'' by the assigned synaptic weights and ``tilted'' by the inputs (Fig.~\ref{fig.hopfield}b,c).

More generally, $W$ need not be symmetric. In biological neural circuits, empirical evidence consistently indicates that cortical neurons have either excitatory or inhibitory synapses, but not both. This fundamental organizing principle governing neural
tissue is known as Dale's law, which implies that if neuron $j$ is excitatory, then $W_{ij}\geq 0$ $\forall i$, and if it is inhibitory, then $W_{ij}\leq 0$ $\forall i$. The matrix $W$ thus has columns of fixed signs and is necessarily asymmetric whenever both excitatory and inhibitory neurons are accounted for. In asymmetric Hopfield networks, an energy function may not exist. This raises the question of whether system \eqref{eq.cthopfield} is guaranteed to asymptotically converge to an equilibrium (and thus reliably retrieve a memory) for any given input $\bm u$. The following theorem establishes sufficient conditions for this property:

\begin{thm}[\cite{forti2002new}] \label{thm.hopfieldglobalstability}
If $W-D$ is Hurwitz diagonally stable, 
then there exists a unique equilibrium point $\bm x^*$ that is globally asymptotically stable for each constant input $\bm u$.
\end{thm}

\begin{rem}
A closely related formulation is the firing rate neural network (FRN) \cite{wilson1972excitatory,betteti2025firing}. The FRN arises from the Hopfield model by treating the membrane potential $\bm x$ as fast variables and directly modeling the firing rates $\bm z = \Phi(\bm x)$ as state variables instead. For $\tau\rightarrow 0$, the potential $\bm x$ in model \eqref{eq.cthopfield} satisfies the quasi-steady state approximation $\bm Dx \approx W\bm z + B\bm u$. Assuming the activation function satisfies the equivariance property $\bm \Phi(D\bm x) = D\bm \Phi(\bm x)$ (as in the case of ReLU), we obtain the nonlinear algebraic equation
\begin{equation}
    D\bm z \approx \Phi(W\bm z + B\bm u).
    \label{eq.frnequilibrium}
\end{equation}
The FRN can be viewed as a first-order relaxation dynamics,
\begin{equation}
    \dot{\bm z} = - D\bm z + \bm \Phi(W\bm z + B\bm u),
    \label{eq.ratedyn}
\end{equation}
whose equilibria $\dot{\bm z}=0$ solve \eqref{eq.frnequilibrium}.
For symmetric $W$, it also admits an energy function analogous to \eqref{eq.hopfield_energy}, leading to an EDM formulation. 
We refer the reader to \cite{bullo2025lecturesneural} for an in-depth comparison of the Hopfield and FRN models.
\end{rem}

\subsection{Continuous-Time Boltzmann Machines}

\begin{figure}
    \centering
    \includegraphics[width=\linewidth]{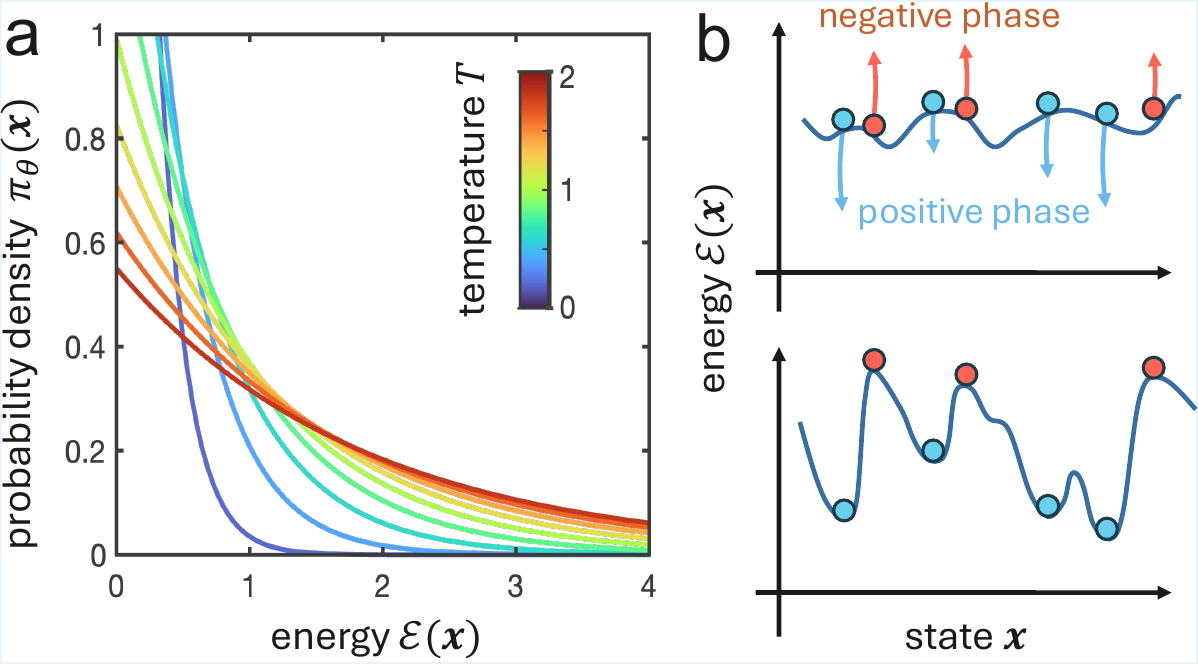}
    \caption{Boltzmann machines for generative modeling and sampling-based inference. (a) Gibbs-Boltzmann distribution, plotted for increasing temperatures. States with higher energy are exponentially less likely to be sampled by the Boltzmann machine. (b) Evolution of the energy landscape via contrastive Hebbian learning. Blue and orange nodes represent states $\bm x$ sampled  from the data distribution $\pi_{\rm data}$ and from the model distribution $\pi_{\theta}$, respectively. Through this contrastive learning, the model reshapes the landscape so that configurations consistent with the data occupy low-energy regions, while inconsistent configurations are pushed to higher energy. Thus, the learned model samples data-like states with higher probability.
    }    
    \label{fig.boltzmann}
\end{figure}

Boltzmann machines extend the Hopfield model to \textit{probabilistic generative modeling} \cite{hinton1983optimal,ackley1985learning}. Rather than simply retrieving pre-specified stored patterns, these machines \textit{learn} an underlying data distribution and can then \textit{generate} new samples from this learned distribution that are statistically consistent with the training data. 
In the EDM framework, a Boltzmann machine can be implemented in continuous time as the stochastic gradient flow \eqref{eq.stogradflow}. The diffusion term $\sqrt{2T}{\rm d}{\bm w}_t$ injects isotropic noise with variance proportional to the temperature $T$. Thus, trajectories no longer monotonically decrease $\mathcal E$ and can escape local minima, enabling broader exploration of the state space.
The key concept in Boltzmann machines is that the energy of a state $\bm x$ determines its probability via the \textit{Gibbs-Boltzmann} distribution: 
\begin{equation}
    \pi_\theta(\bm x)= \frac{1}{Z}\exp(-\frac{\mathcal E(\bm x;\bm\theta)}{T}).
    \label{eq.gibbsdistribution}
\end{equation}
In the discrete case, $\bm x\in\mathcal X$ and $\pi_\theta$ defines the probabilities of each state $x_i$, often described by binary values (i.e., $x_i\in\{\pm 1\}$). In the continuous case, $\bm x\in\mathcal X = \R^N$ and $\pi_\theta$ defines a probability density.
The partition function
\begin{equation}
    Z =
    \begin{cases}
        \sum_{\bm x\in\mathcal{X}} \exp(-\mathcal E(\bm x)/T), &\text{if} \,\, \mathcal X \,\, \text{is discrete}, \\
        \int_{\mathcal X} \exp(-\mathcal E(\bm x)/T) {\rm d}{\bm x}, &\text{if} \,\, \mathcal X \,\, \text{is continuous,}
    \end{cases}
    \label{eq.partition}
\end{equation}
ensures normalization (assuming $Z<\infty$).
The Gibbs-Boltzmann distribution is the canonical equilibrium distribution in statistical mechanics: it expresses state probabilities as functions of energy, with high-energy states being exponentially less likely to occur than low-energy states (Fig.~\ref{fig.boltzmann}a).

Under suitable conditions on $\mathcal E$, the stochastic EDM \eqref{eq.stogradflow} admits $\pi_\theta$ as its unique stationary distribution:

\begin{thm}[\cite{pavliotis2014stochastic}] \label{thm.langevin_gibbs}
Let $\mathcal E\in C^2(\R^N)$ (continuous differentiability), $Z < \infty$ (normalizability), and $\nabla \mathcal E$ be locally Lipschitz (regularity).  
Then, the distribution $\pi_\theta$ is invariant for the system~\eqref{eq.stogradflow}: if $\bm x_0\sim \pi_\theta$, then $\bm x_t\sim \pi_\theta$, $\forall t\ge 0$.

Moreover, suppose $\mathcal E$ satisfies the dissipativity condition
\begin{equation}
    \langle \nabla \mathcal E(\bm x), \bm x \rangle \geq a\norm{\bm x}^2 - b, \,\,\, \forall \norm{x} \geq c,
\end{equation}
for some $a,c>0$ and $b\in\R$. Then, for any initial condition $\bm x_0\in\R^N$, the distribution of $\bm x_t$ converges to $\pi_\theta$ as $t\to\infty$.
\end{thm}

\noindent
Thus, while the deterministic gradient dynamics \eqref{eq.edm} asymptotically converges to local minima of $\mathcal E$, the stochastic dynamics \eqref{eq.stogradflow} asymptotically \emph{samples} from $\pi_\theta$. The temperature $T$ controls the exploration level of the state space. As $T\to 0$, $\pi_\theta$ concentrates on minimizers of $\mathcal E$, and the continuous-time Boltzmann machine reduces to a Hopfield network if $\mathcal E$ is given by Eq.~\eqref{eq.hopfield_energy} and $W$ is symmetric.

\begin{rem}
In practice, the state vector $\bm x$ is partitioned into visible (input/output) variables $\bm v$ and hidden variables $\bm h$. Imposing a bipartite network structure between neurons $\bm v$ and $\bm h$ yields a restricted Boltzmann machine (RBM), which significantly improves the computational tractability for inference and learning \cite{freund1991unsupervised}. A RBM with sufficiently many hidden units can approximate arbitrarily well any probability density over a compact subset of $\R^N$ \cite{le2008representational}.
\end{rem}

\section{Biological versus Machine Learning}
\label{sec.learning}

We begin by reviewing backpropagation\textemdash the canonical training algorithm in machine learning\textemdash and highlighting the fundamental limitations in its interpretation as a biologically plausible mechanism of how the brain learns. We then introduce classical and modern learning mechanisms that address these limitations through local, unsupervised learning rules rather than globally coordinated error signals.

\subsection{Backpropagation and Biological Plausibility}
\label{sec.backprop}

Deep neural networks are typically trained via backpropagation, an algorithm that efficiently computes how each synaptic parameter should change in order to reduce the network's prediction error \cite{rumelhart1986learning}. Backpropagation is a method for \textit{supervised learning}, meaning that the network is trained using labeled input-output pairs: for each input $\bm u$, there is a desired ``target'' output $\bm y_{\rm t}$. 

To formalize it, consider a standard FNN with layers $\ell=0,\ldots,L-1$:
\begin{equation}
    \bm x^{(\ell + 1)} = \bm\Phi \big( \underbrace{ W^{(\ell)} \bm x^{(\ell)} + \bm b^{(\ell)} }_{\bm z^{(\ell)}} \big),
\label{eq.feedforward}
\end{equation}
where $\bm x^{(\ell)}\in\R^{N_\ell}$ is the neural activity, $W^{(\ell)}\in\R^{N_{\ell+1}\times N_\ell}$ is the synaptic matrix, and $\bm b^{(\ell)}\in\R^{N_{\ell+1}}$ is the bias at layer $\ell$. 
The FNN maps an input $\bm x^{(0)} = \bm u$ to an output $\bm x^{(L)} = \bm y$, which is called the \textit{forward} pass.
Training consists of minimizing a loss function $\mathcal L(\bm y,\bm y_{\rm t}; \bm\theta)$ with respect to the trainable parameters $\bm\theta = \{W^{(\ell)}, \bm  b^{(\ell)}\}_{\ell = 0}^{L-1}$. Here, $\mathcal L$ measures the discrepancy between the network output $\bm y$ and the target output $\bm y_{\rm t}$. 
By the chain rule, the gradient of the loss with respect to weights is given by \cite{downing2023gradient}
\begin{equation}
    \pdv{\mathcal L}{W^{(\ell)}_{ij}} = \pdv{\mathcal L}{z_i^{(\ell)}} \pdv{ z^{(\ell)}_i}{W^{(\ell)}_{ij}} = \delta_i^{(\ell)}x_j^{(\ell)}.
\end{equation}
The error signals $\delta_i^{(\ell)}$ are computed \textit{backward} through the network, recursively defined with $\delta_i^{(L)} = \pdv*{\mathcal L}{y_i}$ and
\begin{equation}
    \delta_i^{(\ell)} = \pdv{\mathcal L}{z_i^{(\ell)}} = \Phi'(z_i^{(\ell)}) \sum_{k=1}^{N_{\ell+1}} W_{ki}^{(\ell+1)} \delta_k^{(\ell+1)}.
\end{equation}
Parameters are then updated via (stochastic) gradient descent:
\begin{equation}
    \dot W^{(\ell)}_{ij} = - \eta \pdv{\mathcal L}{W^{(\ell)}_{ij}},
\end{equation}
where $\eta>0$ is the learning rate. Intuitively, a weight $W_{ij}$ changes in proportion to how active the presynaptic neuron $j$ was, and to how much error the postsynaptic neuron $i$ is responsible for.
Note that analogous recursive expressions can be obtained for the biases $\bm b^{(\ell)}$.

Computationally, backpropagation is efficient because all gradients can be computed with a cost comparable to a single forward pass: each error signal $\delta_i^{(\ell)}$ is computed once per neuron and then reused to update all weights $W_{ij}^{(\ell)}$. However, several features of this algorithm raise questions regarding its biological plausibility \cite{lillicrap2020backpropagation}:
\begin{enumerate}
    \item \textit{Nonlocal information}: The update of a synapse $W_{ij}^{(\ell)}$ is based on an error signal $\delta_i^{(\ell+1)}$ that is computed using neural activities and synaptic weights from \textit{all} downstream layers, ultimately reflecting the global output error $\mathcal L(\bm y,\bm y_{\rm t})$.

    \item \textit{Separate backward and forward phases}: 
    Learning requires a separate backward pass after the forward inference pass, while biological neural circuits operate continuously in time without clearly separated phases for learning and inference.
    
    \item \textit{Symmetric connections}: The backward recursion requires multiplication by $(W^{(\ell)})^\transp$, implying symmetric forward and backward synapses for which there is little anatomical evidence
\end{enumerate}

\noindent
These constraints motivate the search for alternative learning mechanisms\textemdash particularly those based on EDMs\textemdash in which learning rules arise from local interactions, recurrent network structures, and operate in an unsupervised manner, instead of relying on explicit global error backpropagation over layered network structures.

\subsection{Hebbian Learning for Deterministic EDMs}
\label{sec.hebbian}

Hebbian learning is named after the psychologist Donald Hebb \cite{hebb2005organization}, who postulated the principle that ``neurons that fire together, wire together.'' Consider a neural network with state $\bm x\in\R^N$, a (possibly recurrent) synaptic matrix $W\in\R^{N\times N}$, and a quadratic energy function 
\begin{equation}
    \mathcal E_H(\bm x;W) = - \frac{1}{2} \bm x^\transp W\bm x.
\label{eq.quadraticenergy}
\end{equation}
Mathematically, Hebbian learning can be interpreted as the gradient descent dynamics on $\mathcal E_H$ \cite{gerstner2002mathematical,munakata2004hebbian,dayan2005theoretical}:
\begin{equation}
    \dot W = - \eta \nabla_{W} \mathcal E_{H} =  \eta \bm x\bm x^\transp,
\label{eq.hebbian}
\end{equation}
\noindent
where $\eta>0$. Fig.~\ref{fig.hopfield}d shows the energy landscape evolving under this learning rule. The rule is said to be \textit{cooperative} because a synapse $W_{ij}$, which transmits signals from the presynaptic neuron $j$ to the postsynaptic neuron $i$, increases whenever both neurons are simultaneously active. It is also \textit{local} since the learning dynamics  \eqref{eq.paramlearning} depend only on first-neighbor states in the network, i.e., $\dot W_{ij} = g(W_{ij};x_i,x_j)$. And it is \textit{unsupervised} since the update depends only on neural activity and not on external target signals.

The following theorem formalizes how Hebbian learning can be used to assign the synaptic weights $W_{ij}$ such that prescribed memory patterns become attractors of the Hopfield network \eqref{eq.cthopfield}. More generally, this result applies to any EDM (partially) governed by the quadratic energy function~\eqref{eq.quadraticenergy}.

\begin{thm} \label{thm.hebbian}
    Let $\bm\xi^{(\mu)}\in\R^N$, for $\mu=1,\ldots,K$, denote the patterns to be stored in the energy function \eqref{eq.quadraticenergy}. 
    The expectation over the pattern distribution is
    \begin{equation}
        \mathbb{E}[\dot W] = \eta \mathbb{E}[\bm x\bm x^\transp] = \frac{\eta}{K} \sum_{\mu=1}^K \bm\xi^{(\mu)} (\bm\xi^{(\mu)})^\transp. 
    \end{equation}
    Thus, the learning dynamics \eqref{eq.hebbian} converge asymptotically to
    \begin{equation}
        W\propto \sum_{\mu=1}^K \bm\xi^{(\mu)} (\bm\xi^{(\mu)})^\transp. 
    \label{eq.hebbianasymptotic}
    \end{equation}
\end{thm}

\noindent

Oja's rule \cite{oja1982simplified} adapted Eq.~\eqref{eq.hebbian} by introducing a quadratic term,
\begin{equation}
    \dot W_{ij} = \eta (x_i x_j - x_i^2 W_{ij}),
\label{eq.oja}
\end{equation}
which implements a form of \textit{synaptic depression}. The second term counteracts unbounded weight growth by allowing synaptic weights to decay. The resulting stabilization of the learning dynamics is formalized next.

\begin{thm} \label{thm.ojapca}
Consider a single linear neuron with output $y(t)=\bm w(t)^\transp \bm x(t)$, whose synaptic weights $\bm w\in\R^{N}$ evolve according to Oja's rule $\dot{\bm w}=\eta(y\bm x-y^2\bm w)$. Then, the expected weight dynamics satisfy
\begin{equation}
\mathbb{E}[\dot{\bm w}] = \eta \left( C\bm w - (\bm w^\transp C\bm w)\bm w \right),
\label{eq.ojaexpected}
\end{equation}
where $C = \mathbb E[\bm x\bm x^\transp]$ is the covariance matrix. 
Moreover,
\begin{equation}
\label{eq.ojanorm}
    \mathbb{E}\left[\dv{}{t} \norm{\bm w}^2\right] = 2\eta (\bm w^\top C\bm w) (1-\norm{\bm w}^2), 
\end{equation}
so the unit sphere $\|\bm w\|=1$ is an invariant manifold.
\end{thm}

\begin{rem}
Oja's rule can be interpreted as an online, biologically plausible implementation of principal component analysis (PCA) \cite{oja1982simplified,oja1989neural}. Let the eigenvalues of $C$ satisfy $\lambda_1>\lambda_2\ge\cdots\ge\lambda_N\ge 0$, with corresponding orthonormal eigenvectors $\{\bm v_k\}_{k=1}^N$ (i.e., the principal components). 
Eq.~\eqref{eq.ojanorm} shows that trajectories asymptotically converge toward the unit sphere $\|\bm w\|=1$ for $\bm w(0)\neq 0$. On this sphere, the equilibria satisfy $\lambda = {\bm w^*}^\transp C \bm w^*$, implying that $\bm w^* = \pm \bm v_k$. Moreover, linearization of the mean dynamics \eqref{eq.ojaexpected} shows that $\pm \bm v_1$ are the only asymptotically stable equilibria, whereas $\pm \bm v_k$, $k\geq 2$, are saddle points. Therefore, under the mean dynamics, Oja's rule drives the weight vector toward the first principal component of the covariance matrix.
\end{rem}

It is important to note that there are many widely used local plasticity rules that are different from Oja's rule and demonstrate competitive performance on various computational problems \cite{bienenstock1982theory, krotov2019unsupervised, pehlevan2015hebbian, grinberg2019local, kozachkov2020achieving}. 


\subsection{Contrastive Hebbian Learning for Stochastic EDMs}

Consider the Gibbs-Boltzmann distribution \eqref{eq.gibbsdistribution}, parameterized by $\bm\theta$, and the distribution of training samples $\bm x \sim \pi_{\rm data}$. Contrastive Hebbian learning (CHL) \cite{ackley1985learning} seeks parameters $\bm\theta$ that minimize the Kullback-Leibler (KL) divergence between the model and data distributions, given by
\begin{equation}
    D_{\rm KL}(\pi_{\rm data} \| \pi_\theta) = \mathbb{E}_{\pi_{\rm data}} \left[\log \frac{\pi_{\rm data}(\bm x)}{\pi_\theta(\bm x)} \right].
\end{equation}
Here, $\mathbb{E}_{\pi}[\bm F(\bm x)] = \int \bm F(\bm x) \pi(\bm x){\rm d}\bm x$ denotes the expectation of a function $\bm F$ under a distribution $\pi$.
Since $\pi_{\rm data}$ does not depend on $\bm\theta$, minimizing $D_{\rm KL}$ is equivalent to minimizing the negative log-likelihood $\mathcal L(\bm\theta) = - \mathbb{E}_{\pi_{\rm data}}[\log \pi_\theta(\bm x)]$ with respect to the parameters $\bm\theta$. 
For the quadratic energy function \eqref{eq.quadraticenergy}, with parameters $\bm\theta = \{W_{ij}\}_{i,j=1}^N$, the weight update rule is thus given by the gradient-descent dynamics:
\begin{equation}
    \dot W_{ij} = - \eta \pdv{\mathcal L(\bm\theta)}{W_{ij}} = \eta\big( \underbrace{\mathbb E_{\pi_{\rm data}} [x_ix_j]}_{\text{positive phase}} - \underbrace{\mathbb E_{\pi_{\theta}} [x_ix_j]}_{\text{negative phase}} \big).
\end{equation}

Learning is again unsupervised and correlation-based, but now \textit{contrastive}. As shown in Fig.~\ref{fig.boltzmann}b, the positive phase lowers the energy of observed patterns in the training data (increasing their probability), while the negative phase raises the energy of patterns sampled by the model (decreasing their probability) \cite{downing2023gradient}. At convergence, a Boltzmann machine trained with CHL minimizes the KL divergence between the model and data distributions, so that $\pi_\theta$ approaches $\pi_{\rm data}$. 

\subsection{Equilibrium Propagation}

Equilibrium propagation (EqProp) \cite{scellier2017equilibrium} is a learning mechanism designed to bridge EDMs and gradient-based learning. Unlike backpropagation, EqProp computes gradients through the continuous-time relaxation of an EDM, using only local interactions and eliminating the need for a separate backward pass. This property makes it particularly suitable for direct analog implementations \cite{stern2021supervised}.

EqProp performs supervised learning. Let $\bm y_{\rm t}\in\R^q$ be the target output and $\bm y\in\R^q$ be the network output, which is an algebraic function of the neural activities:
\begin{equation}
    \bm y = \bm H(\bm x;\bm\theta,\bm u),
\end{equation}
\noindent
where $\bm H : \R^N \mapsto \R^q$.
As in Sec.~\ref{sec.backprop}, the supervised loss is given by $\mathcal L(\bm y,\bm y_{\rm t})$. In the EDM framework, given a fixed input $\bm u$, the network output is evaluated only after the EDM \eqref{eq.edm} relaxes to the equilibrium state $\bm x^*$. Thus, the trainable parameters $\bm\theta$ only affect the loss $\mathcal L$ through the equilibrium state, given by $\bm y^* = \bm H(\bm x^*;\bm\theta,\bm u)$. As a result, the goal becomes instead to minimize the objective function $\mathcal J(\bm\theta) = \mathcal L(\bm y^*,\bm y_{\rm t})$ with respect to $\bm \theta$. 

The parameter learning process of EqProp consists of two phases, which are iterated repeatedly. During the \textit{free phase}, the EDM \eqref{eq.edm} evolves under the original energy $\mathcal E$ for some input $\bm u$ and relaxes to an equilibrium $\bm x^{(0)}$ satisfying $\nabla_{\bm x} \mathcal E(\bm x^{(0)};\bm\theta,\bm u)=0$. Afterwards, during the \textit{nudged phase}, the energy function $\mathcal E$ is slightly perturbed by adding a small term proportional to the loss:
\begin{equation}
    \mathcal E_\beta(\bm x;\bm\theta,\bm u,\bm y_{\rm t}) = \mathcal E(\bm x;\bm\theta,\bm u) + \beta \mathcal L(\bm y,\bm y_{\rm t}),
    \label{eq.augmented_energy}
\end{equation}
where $\beta>0$ is a small ``nudging'' parameter. The EDM becomes
\begin{equation}
    \dot{\bm x} = -\nabla_{\bm x} \mathcal E_\beta(\bm x) = - \nabla_{\bm x} \mathcal E(\bm x) - \beta\nabla_{\bm x} \mathcal L(\bm y,\bm y_{\rm t}),
\end{equation}
for which the system relaxes to a new equilibrium $\bm x^{(\beta)}$ satisfying $\nabla_{\bm x} \mathcal E_\beta(\bm x^{(\beta)})=0$. The small displacement $\bm x^{(\beta)} - \bm x^{(0)}$ encodes an error signal, measuring how the equilibrium shifts as a function of the current loss. Crucially, this shift contains exactly the information required to compute the gradient of $\mathcal J$, as formalized in the next theorem. No backward pass is required, given that the system dynamics transport the error.

\begin{thm}[\cite{scellier2017equilibrium}] \label{thm.eqprop}
Let $\bm x^{(0)}$ be a nondegenerate stable equilibrium. Then, the gradient of the objective function is
\begin{equation}
    \nabla_{\bm\theta} \mathcal J(\bm\theta) = \lim_{\beta\to 0} \frac{1}{\beta} \left( \nabla_{\bm\theta} \mathcal E(\bm x^{(\beta)}) - \nabla_{\bm\theta} \mathcal E(\bm x^{(0)}) \right),
\label{eq.eqpropgradient}
\end{equation}
and the resulting EqProp learning rule is $\dot{\bm\theta} = -\eta\nabla\mathcal J(\bm\theta)$.
\end{thm}

\begin{cor}[\cite{scellier2017equilibrium}] \label{cor.eqprop2chl}
If $\mathcal E=\mathcal E_H$, EqProp reduces to CHL:
\begin{equation}
    \dot W_{ij} = \frac{\eta}{\beta} \big(x_i^{(\beta)} x_j^{(\beta)} - x_i^{(0)} x_j^{(0)} \big).
\end{equation}
\end{cor}

Backpropagation, CHL, and EqProp differ less in \textit{what} they optimize than in \textit{how} gradients are computed and transported. Backpropagation evaluates gradients explicitly through dedicated backward passes; CHL evaluates gradients through correlations among data and model samples; and EqProp extracts the gradients implicitly through dynamical relaxation. 
In the context of supervised learning, Corollary~\ref{cor.eqprop2chl} shows that EqProp reduces to CHL, which in turn is equivalent to backpropagation under certain conditions (e.g., models with linear outputs \cite{xie2003equivalence} or no hidden units \cite{movellan1991contrastive}). EqProp has also been proven equivalent to recurrent backpropagation, used for training RNNs \cite{scellier2019equivalence}.
These mechanisms help close the gap between biological and machine learning.

\section{Dense Associative Memory}
\label{sec.dam}

Dense Associative Memory (DenseAM) is a generalization of the celebrated Hopfield networks \cite{krotov2016dense,krotov2018dense}. While Hopfield networks are elegant mathematical models, they are known to have a very small \textit{information storage capacity}: the number of patterns that can be reliably stored and retrieved scales poorly with the network size. As a result, the storage capacity of Hopfield networks is insufficient for practical AI applications. DenseAMs are specifically designed to retain all of the benefits of Hopfield networks, but rectify their small information storage issue. 

DenseAMs can be formulated with discrete or continuous variables, and in discrete or continuous time. First, we focus on DenseAMs with discrete states and discrete asynchronous updates. Consider a set of discrete variables $\sigma_i \in \{\pm 1\}$, for $i=1,...,N$. 
We refer to the individual elements of the state vector $\bm\sigma=(\sigma_1,\ldots,\sigma_N)^\transp$ as neurons or spins. In addition, let the model have $K$ memory vectors $\boldsymbol{\xi}^{(\mu)}\in\{\pm 1\}^N$.

The energy function of a DenseAM is defined as 
\begin{equation}
    \mathcal E(\bm\sigma) = - \sum\limits_{\mu=1}^K F\Big((\bm\xi^{(\mu)})^\transp \bm\sigma\Big).
    \label{eq: DenseAM energy}
\end{equation}
Thus, the energy is a finite sum of smooth functions $F$ that depend on the finite number of discrete variables. Assuming $F$ has no singularities (i.e., it does not take infinite values for finite arguments), the energy is finite and lower bounded.
The goal of this model is to start at some initial state $\bm\sigma(0)$, which typically corresponds to a high-energy state, and lower the energy \eqref{eq: DenseAM energy} by flipping the elements of the state vector. The dynamics of flipping stop when no further single-element flip can reduce the energy. At that point, the network has reached a local minimum of the energy. 
This update rule is typically written as a discrete-time dynamical system:
\begin{equation}
    \sigma_i{(t+1)} =\text{sign}\bigg[\sum\limits_{\mu=1}^K \xi^{(\mu)}_i \Phi \Big(\sum\limits_{j\neq i}^N \xi^{(\mu)}_j \sigma_j(t)\Big) \bigg] ,
    \label{eq: DenseAM update simplified}
\end{equation}
where here we introduce the activation function $\Phi(\cdot) = F^\prime(\cdot)$, which is a derivative of the function $F$ defining the energy \cite{krotov2025modern}.  
The dynamical system \eqref{eq: DenseAM update simplified} decreases the energy value at each iteration (see \cite{krotov2025modern} for a simple proof). Thus, if we repeatedly apply these update rules to the state vector, the system will asymptotically reach a steady state\textemdash no single neuron flip can further reduce the energy. DenseAMs therefore constitute a class of EDMs.

\subsection{Storage Capacity}
How many memories or local minima can such a system store and successfully retrieve? The DenseAM, specified by \eqref{eq: DenseAM energy}, can be defined for any number $K$ of memories. However, if too many memory vectors are packed inside the $N$-dimensional discrete space, the local minima of the energy will no longer correspond to the stored patterns. 
\textit{Storage capacity} thus measures the maximum number of patterns $K^{\rm max}$ that can be reliably stored as attractors of an $N$-dimensional DenseAM. 
%
%
In general, $K^\text{max}$ depends on the specific structure of the stored memories. We derive a statistical scaling law for this memory capacity assuming that the patterns are uniformly drawn at random:
\begin{equation} \label{eq.randsamp}
    \xi^{(\mu)}_i = \begin{cases}
            +1, \ \ \text{with probability}\  \frac{1}{2}, \\
            -1, \ \ \text{with probability}\  \frac{1}{2}.
    \end{cases}
\end{equation}
With this distribution, it is easy to compute the correlation functions for these variables. The one-point and two-point correlation functions are, respectively, $\langle\xi^{(\mu)}_i \rangle = 0$ and $\langle \xi^{(\mu)}_i \xi^{(\nu)}_j \rangle = \delta_{\mu \nu} \delta_{i j}$,  where $\delta_{ij}$ denotes the Kronecker delta.

To quantify the information storage capacity of this model, we use the following trick. First, we initialize the model in a state corresponding to one of the memories, say $\bm\sigma(0) = \bm\xi^{(1)}$, and let it evolve in time according to the update rule. If the pattern $\bm\xi^{(1)}$ corresponds to a local minimum, then that state must be stable. In other words, the dynamics should not change that initial state. Mathematically, this means that the updated state 
\begin{equation} \label{eq:stability analysis}
{\small
\begin{split}
    \sigma_i{(t\hspace{-2pt}+\hspace{-2pt}1)} & =\text{sign}\bigg[ \xi^{(1)}_i \Phi \Big(\sum\limits_{j\neq i}^N \xi^{(1)}_j \xi^{(1)}_j\Big) \hspace{-1pt} + \sum\limits_{\mu=2}^K \xi^{(\mu)}_i \Phi \Big(\sum\limits_{j\neq i}^N \xi^{(\mu)}_j \xi^{(1)}_j\Big) \bigg]  \\
    & = \text{sign}\bigg[ \underbrace{\xi^{(1)}_i \Phi (N-1)}_{\rm signal} + \underbrace{\sum\limits_{\mu=2}^K \xi^{(\mu)}_i \Phi \Big(\sum\limits_{j\neq i}^N \xi^{(\mu)}_j\ \xi^{(1)}_j\Big)}_{\rm noise} \bigg]
\end{split}
}
\end{equation}

\noindent
should coincide with $\xi^{(1)}_i$.

Assuming $\Phi$ is a non-negative function, the \textit{signal} term pushes the argument of the sign function towards alignment with the desired pattern $\xi^{(1)}_i$. The \textit{noise} term generally pushes that argument away from the desired pattern and, in some situations, may outweigh the signal term. Below, we compute the characteristic magnitude of the noise term and determine when it becomes dominant and destroys the stability of the target memory. Specifically, we can compute the mean and variance of the noise term. The mean is given by
\begin{equation}
    \langle{\rm noise}\rangle = \bigg\langle \sum\limits_{\mu=2}^K \xi^{(\mu)}_i\ \Phi \Big(\sum\limits_{j\neq i}^N \xi^{(\mu)}_j \xi^{(1)}_j\Big) \bigg\rangle = 0, 
\end{equation}
since the index $i$ appears only once in the correlator, and the variance is given by \cite{krotov2025modern}
\begin{equation} 
    \langle{\rm noise}^2 \rangle  = (K-1) \Big\langle \Phi \Big(\sum\limits_{j\neq i}^N \xi^{(\mu)}_j\Big)^2 \Big\rangle.
\end{equation}

Now, we restrict our calculation to the class of \textit{power energy functions} so that
\begin{equation} \label{eq.powerenergy}
    F(\cdot) = \frac{1}{n}(\cdot)^n,  \ \ \Phi(\cdot) = (\cdot)^{n-1},  \ \ \text{where} \ n \ \text{is an integer} .
\end{equation}
In this case, the variance of the noise can be computed exactly (through the generating function) \cite{chaudhry2023long,krotov2025modern}: 
\begin{equation}
    \Sigma^2 = \langle {\rm noise}^2 \rangle = (2n-3)!! K N^{n-1} \, .
\end{equation}
We are then ready to compute the probability of an error in the memory retrieval. The noise term in \eqref{eq:stability analysis} is a sum over many independent random variables. When $K$ and $N$ are large, this noise term behaves approximately as a Gaussian random variable. When the noise has the same sign as the signal, the noise term pushes the update in the right direction and does not cause issues. The problem arises when the noise is large and its sign is opposite to that of the signal. In this case, the noise can outweigh the signal and unintentionally flip the state of a neuron of interest, creating a \textit{bit error}. The probability of this event is given by the area under the tail of a Gaussian distribution:
\begin{equation}
\begin{split}
    \mathbb P\big[{\rm bit \,\, error }\big] &:= \mathbb P\big[{\rm signal} < {\rm noise}\big] \\
    &= \int_{\Phi(N-1)}^\infty \frac{1}{\sqrt{2\pi\Sigma^2}}e^{-\frac{x^2}{2\Sigma^2}} {\rm d}x \\ 
    &= \int_{\frac{\Phi(N-1)}{\Sigma}}^\infty \frac{1}{\sqrt{2\pi}}e^{-\frac{y^2}{2}} {\rm d}y \\
    &=: g\Big(\frac{\Phi(N-1)}{\Sigma}\Big). 
\end{split}
\end{equation}
The storage capacity of a model is thus formally defined as:
\begin{defin} \label{def.smallerrorcap}
Small-error capacity is the largest number of patterns, $K\leq K^{\rm max}$, such that
\begin{equation}
    \mathbb P\big[{\rm bit \,\, error }\big] < \epsilon,
\end{equation}
where $0<\epsilon\leq 1$ defines the error tolerance.
\end{defin}

For the DenseAMs, if the probability of error is sought to be smaller than a certain value $\epsilon$, then the following inequality must be satisfied:
\begin{equation}
    \Phi(N-1)> \alpha \Sigma , 
\end{equation}
where $\alpha$ is a numerical constant independent of $K$, $N$, and $n$ (for $\epsilon = 1\%$ error, we have that $\alpha\approx 2.576$). This translates into the following capacity:
\begin{equation}
    K<K^{\max} = \frac{1}{\alpha^2 (2n-3)!!} N^{n-1} \, .
    \label{eq: capacity bound}
\end{equation}
Thus, as long as the number of memories is smaller than $K^{\max}$, a DenseAM initialized in one of the memories remains there and the dynamics do not steer away from it. It turns out that this is precisely the point when associative memory retrieval breaks. If the number of memories is smaller than $K^{\max}$, the model works as intended. Once $K$ exceeds $K^{\max}$, reliable retrieval breaks. However, this does not mean that the model becomes useless in that regime; in fact, it instead becomes a generative model \cite{pham2025memorization}.

\begin{rem}
    It is also possible to determine the storage capacity in scenarios in which almost no errors are allowed (instead of a small percentage of errors). In this case, the event probability of a single-bit error should be smaller than $\epsilon = 1/N$, resulting in  $K^{\max}_\text{no errors} \sim \frac{N^{n-1}}{\log(N)}$ for $N\rightarrow\infty$ \cite{krotov2016dense}.
\end{rem}

\subsection{Limiting Cases}
It is instructive to study a few limiting cases of the general DenseAM family (\ref{eq: DenseAM energy}). Each of these models is frequently studied in the literature and has distinct properties.

\smallskip
\subsubsection{The Hopfield Model ($n=2$)} The simplest, and the most popular, example of DenseAMs is the Hopfield model. One can obtain it from the general form (\ref{eq: DenseAM energy}) choosing $F(\cdot) = \frac{1}{2}(\cdot)^2$. The energy function can be written as 
\begin{equation}
\begin{split}
    \mathcal E(\bm\sigma) = - \frac{1}{2} \sum\limits_{\mu=1}^K \Big(\sum\limits_{i=1}^N \xi^{(\mu)}_i \sigma_i\Big)^2 = - \frac{1}{2} \sum\limits_{i,j=1}^N \sigma_i T_{ij} \sigma_j,
\end{split}
\end{equation}
\noindent
where $T_{ij} = \sum_{\mu=1}^K \xi^{(\mu)}_i \xi^{(\mu)}_j$. 
In this case, according to the general result (\ref{eq: capacity bound}), the storage capacity scales linearly with the size of the network:
\begin{equation}\label{eq.hopfieldlaw}
    K^{\max} \sim N.
\end{equation}
This is the famous $K^{\max} \approx 0.14 N$ scaling law from Hopfield's paper \cite{hopfield1982neural}. Later, it was also derived using tools from statistical mechanics \cite{amit1985storing}. 
This linear scaling presents a major practical limitation. In the end, the hallmark of modern AI is the ability to store and process large amounts of information, a property severely limited by relation~\eqref{eq.hopfieldlaw}.

\smallskip
\subsubsection{DenseAM with $n=3$} Fortunately, the limited scaling law disappears in DenseAMs for a more rapidly peaking energy function (obtained via an alternative activation function). For $F(\cdot) = \frac{1}{3}(\cdot)^3$, the energy is given by 
\begin{equation}
\begin{split}
    \mathcal E(\bm\sigma) = - \frac{1}{3} \sum\limits_{\mu=1}^K \Big(\sum\limits_{i=1}^N \xi^{(\mu)}_i \sigma_i\Big)^3 = - \frac{1}{3} \sum\limits_{i,j,k=1}^N  T_{ijk} \sigma_i \sigma_j \sigma_k,
\end{split}
\end{equation}
where $T_{ijk} = \sum_{\mu=1}^K \xi^{(\mu)}_i \xi^{(\mu)}_j \xi^{(\mu)}_k$. The storage capacity thus scales as
\begin{equation}
    K^{\max} \sim N^2 ,
\end{equation}
which is significantly faster than linearly. 

\smallskip
\subsubsection{DenseAM with $F(\cdot) = \exp(\cdot)$} It turns out DenseAMs can even achieve exponentially large memory storage capacity. For the exponential function $F(\cdot) = \exp(\cdot)$ \cite{demircigil2017model,lucibello2024exponential}, the number of memories that this DenseAM can store and retrieve scales as  
\begin{equation}
    K^{\max} \sim 2^{\frac{N}{2}}, 
\end{equation}
which is more than sufficient for storing any practically relevant amount of information. Note that this number is the square root of the total number of binary patterns of the network (i.e., all possible $N$-dimensional binary vectors). Despite its huge memory storage capacity, this model retains strong error correcting capabilities and has large basins of attraction around each stored memory. 

\subsection{General Dense Associative Memory with Binary States}
Although simple models represented by (\ref{eq: DenseAM energy}) illustrate the computational capabilities of DenseAMs,  more general energy functions are frequently studied. For binary DenseAM models, the general form of the energy function is given by
\begin{equation}
   \mathcal E = - Q\Big[ \sum\limits_{\mu=1}^K F\Big( S\big[ \boldsymbol{\xi}^{(\mu)}, \boldsymbol{\sigma}\big]\Big) \Big],\label{DenseAM energy}
\end{equation}
where the function $F$ is a rapidly growing separation function (e.g., power or exponential), $S[\boldsymbol{x}, \boldsymbol{x'}]$ is a similarity function (e.g., a dot product or a Euclidean distance), and $Q$ is a scalar monotonic function (e.g., linear or logarithmic). There are many possible combinations of various functions $F(\cdot), S(\cdot,\cdot)$, and $Q(\cdot)$ that lead to different models from the DenseAM family \cite{krotov2016dense,demircigil2017model,ramsauer2020hopfield,krotov2021large, millidge2022universal,burns2022simplicial,saha23end}.

\subsection{Continuous-Time Dense Associative Memory Models}
To train these networks with the backpropagation algorithm, it is helpful to use continuous state variables. In this case, the DenseAM dynamics can be described by the continuous-time dynamical system \cite{krotov2021large}:
\begin{equation}\label{eq:Krotov-Hopfield diff eqs}
\left\{
\begin{aligned}
\tau_v \dot v_i &= \sum_{\mu=1}^{N_h} \xi_{\mu i} f_\mu  - v_i,\\
\tau_h \dot h_\mu &= \sum_{i=1}^{N_v} \xi_{\mu i} g_i - h_\mu.
\end{aligned}
\right.
\end{equation}
This network describes a temporal evolution of two groups of neurons: $N_v$ visible neurons $v_i$ coupled to $N_h$ hidden neurons $h_i$. The nonlinearities here are the functions $f_\mu$ and $g_i$, which can be interpreted as neural activation functions or firing rates. The hidden units can be integrated out from model \eqref{eq:Krotov-Hopfield diff eqs} to obtain an effective theory on visible neurons only. Under a certain choice of the activation functions, this procedure results in models with superlinear memory storage capacity as a function of the number of visible neurons. There are also models in which the capacity is superlinear in the number of hidden units  \cite{kafraj2026biologically, hoover2025dense}. 

There are a variety of models belonging to the DenseAM family. Some utilize hierarchical representations of memories \cite{krotov2021hierarchical, hoover2022a}, others represent the state vector as a collection of tokens that evolve in time to describe transformers \cite{hoover2024energy}. There also exist generalizations with multiple energy functions for the system \cite{dehmamy2025nrgpt}, as well as biological models for non-neuronal cells in the brain called astrocytes \cite{kozachkov2025neuron}. Finally, there exist DenseAM implementations where the dynamics of the physical system itself can perform inference, such as in analog realizations with integrated circuits \cite{bacvanski2025dense}.  

On the AI application front, DenseAMs have been successfully applied to various domains, including vision and graphs \cite{hoover2024energy, liang2022modern}, physical simulations \cite{zhang2025operator}, computational biology \cite{widrich2020modern}, language modeling \cite{dehmamy2025nrgpt}, and many others. The combination of large information storage capabilities with the flexibility to incorporate various inductive biases makes the DenseAM family of models appealing for many tasks in AI, neurobiology, and neuromorphic computing \cite{krotov2023new}.

\section{Oscillator-Based Models for Learning and Optimization}
\label{sec.oscillator}

The neurocomputational models discussed thus far rely on a neural interpretation that contradicts biological evidence: at equilibrium, neurons settle at constant activity levels and remain persistently active. 
In contrast, biological neural systems rarely operate at static membrane potentials. Cortical circuits exhibit rhythmic activity, transient synchronization, and periodic dynamics.
A natural generalization is to replace equilibrium attractors with limit-cycle attractors, where information is encoded in the relative phases of coupled oscillators. This formulation offers key advantages in storage capacity (Sec.~\ref{sec.capacity}) and physical implementation (Sec. \ref{sec.oim}).

\begin{figure*}
    \centering
    \includegraphics[width=\linewidth]{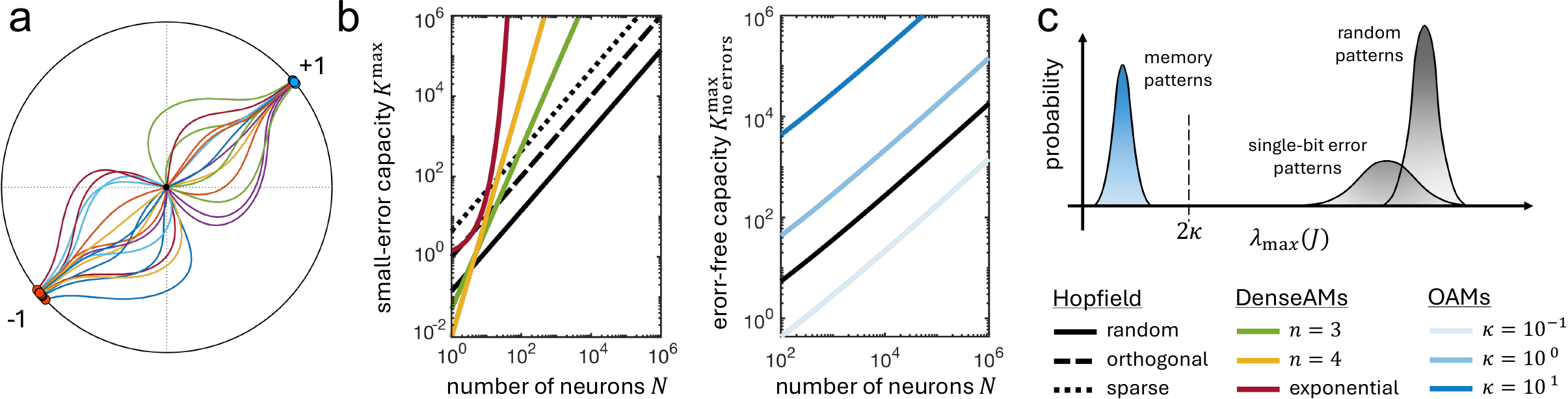}
    \caption{Oscillatory associative memory models. 
    (a) Binary memory patterns are encoded as phase-locked configurations, which form stable equilibria due to the second-harmonic coupling in OAMs. 
    (b) Capacity scaling for small-error (left) and error-free (right) retrieval. The curves are shown for Hopfield networks (assuming binary random, orthogonal, and sparse patterns), DAMs (with different functions $F$), and OAMs (with different coupling strengths $\kappa$). 
    (c) Stability diagram of the OAM, showing the distribution of $\lambda_{\rm max}(J(\bm\xi^{(\mu)}))$ for stored binary memory patterns, single-bit perturbations of those patterns, and random binary configurations. For a suitable choice of $\kappa$, the desired memory patterns can be made stable while all other (undesired) configurations become unstable, thereby eliminating spurious attractors.}
    \label{fig.oam}
\end{figure*} 

Let $\bm\phi\in\mathbb S^N$ denote the oscillator phases. An oscillatory associative memory model (OAM) is given by \cite{nishikawa2004capacity}
\begin{equation}
    \dot{\phi_i} = \omega + \sum_{j=1}^N W_{ij}\sin(\Phi_{ij}) + \frac{\kappa}{N}\sum_{j=1}^N \sin(2\Phi_{ij}),
\label{eq.oam}
\end{equation}
where $\Phi_{ij} = \phi_j - \phi_i$, $\omega$ is the natural frequency, $W\in\R^{N\times N}$ is the synaptic matrix, and $\kappa\geq 0$ is the global coupling strength. When the second-harmonic coupling is neglected (i.e., $\kappa=0$), the OAM reduces to the well-studied network of Kuramoto oscillators \cite{Dorfler2014}.

Because all natural frequencies are identical, the change of variable $\phi_i \mapsto \phi_i + \omega t$ transforms \eqref{eq.oam} into a co-rotating frame, so that limit cycles correspond to equilibrium points. If $W=W^\transp$, the OAM can then be represented as the EDM $\dot{\bm\phi}=-\nabla \mathcal E(\bm\phi)$, with the following energy function:
\begin{equation}
    \mathcal E(\bm\phi) = - \frac{1}{2} \sum_{i,j=1}^N  W_{ij}\cos(\Phi_{ij}) - \frac{\kappa}{4N} \sum_{i,j=1}^N \cos(2\Phi_{ij}).
\label{eq.oamenergy}
\end{equation}
which satisfies $\dot{\mathcal E}(\bm\phi)\leq 0$, $\forall\bm\phi$. Let $\bm\xi^{(\mu)}\in \{\pm 1\}^N$ be a \textit{binary} memory pattern sought to be stored. A \textit{phase-locked} equilibrium $\bm\phi^*$ encodes a memory pattern $\bm\xi^{(\mu)}$ if 
\begin{equation}
    |\phi_j^* - \phi_i^*| =  \begin{cases}
        0, \,\, &\text{if} \,\, \xi^{(\mu)}_i = \xi^{(\mu)}_j, \\
        \pi, \,\, &\text{if} \,\, \xi^{(\mu)}_i \neq \xi^{(\mu)}_j.
    \end{cases}
\label{eq.oamphaselock}
\end{equation}
Thus, neurons representing the same (opposite) binary value synchronize (anti-synchronize), as shown in
Fig.~\ref{fig.oam}a. The second-harmonic term in \eqref{eq.oamenergy} is minimized exactly at such configurations, implying the existence of $2^N$ equilibria corresponding to every possible binary pattern (up to rotational symmetry).
Since the energy \eqref{eq.oamenergy} has the same pairwise structure as \eqref{eq.quadraticenergy}, the Hebbian rule \eqref{eq.hebbianasymptotic} can be directly applied as a learning mechanism for the weights $W_{ij}$ of an OAM.

\begin{rem}
    Richer alphabets can be encoded by generalized OAMs with higher-order harmonics \cite{follmann2014phase,nagerl2025higher}. For example, adding $\frac{\kappa}{N}\sum_j \sin(4\Phi_{ij})$ to \eqref{eq.oam} introduces stable phase differences $\{0,\pi/2,\pi,3\pi/2\}$, encoding quaternary patterns.
\end{rem}

\subsection{Storage Capacity with Error-Free Retrieval}
\label{sec.capacity}

We recall that storage capacity is the largest number of patterns that can be \textit{reliably} stored as attractors of the EDM. In the literature, this notion can be quantified under two distinct criteria depending on what it means for a (typically binary) pattern to be reliably retrieved:
\begin{enumerate}
    \item[i)] \textit{Small-error} capacity: Retrieval is considered successful if the fraction of neurons for which the retrieved state $\bm x^*$ differs from the specified pattern $\bm\xi^{(\mu)}$ is small. (Definition~\ref{def.smallerrorcap}). Occasional bit errors $x_i^*\neq\xi^{(\mu)}_i$ are therefore permitted.

    \item[ii)] \textit{Error-free} capacity: Retrieval is defined more conservatively, as each stored pattern $\bm\xi^{(\mu)}$ must correspond exactly to a strict local minimum of $\mathcal E$ and hence to a stable equilibrium $\bm x^*$ of the EDM  (Definition~\ref{def.errorfreecap}). 
\end{enumerate}
In the error-free definition, retrieval must converge exactly to the stored pattern rather than to a nearby configuration. However, as the number of stored patterns increases, competition among their basins of attraction becomes unavoidable. Beyond capacity, the energy landscape becomes increasingly rugged: some stored patterns cease to be local minima, while additional local minima may emerge at intermediate points that do not coincide with any pattern $\bm \xi^{(\mu)}$ of interest. These undesired minima are known as \textit{spurious attractors} (Fig.~\ref{fig.hopfield}b), which are a major bottleneck of Hopfield models. Here, we analyze both the small-error and error-free capacities, showing that OAMs\textemdash when properly tuned\textemdash can mitigate the existence of spurious attractors. 

We now formalize the notion of error-free capacity.
By the Hebbian rule \eqref{eq.hebbianasymptotic}, we have that $W = \frac{1}{N} \Xi\Xi^\transp$, where $\Xi = [\bm\xi^{(1)} \, \ldots \, \bm\xi^{(K)}]$. Let $\bm\xi^{(\mu)}$ be i.i.d. random variables sampled according to \eqref{eq.randsamp}, so that $\Xi$ is a random matrix. The event that an EDM, trained via Hebbian rule, correctly stores a pattern $\bm\xi^{(\mu)}$ is denoted by 
\begin{equation*}
    \mathcal A(\mu) = \left\{ \Xi\in\{\pm 1\}^{N\times K} : \exists \, \text{stable} \, \bm x^* \, \text{s.t.} \, \Gamma(\bm x^*) = \bm\xi^{(\mu)} \right\}.
\end{equation*}
Here, $\Gamma$ is a prespecified decoding map from the EDM's continuous state space to the discrete pattern space. An equilibrium $\bm x^*$ is said to store $\bm\xi^{(\mu)}$ if it is stable and maps back to $\bm\xi^{(\mu)}$ under $\Gamma$. In general, $\Gamma$ may be many-to-one as multiple equilibria can represent the same pattern. For Hopfield networks, a common choice is $\Gamma(\cdot) = \operatorname{sign}(\cdot)$, so that storage implies that $\bm x^*$ and $\bm\xi^{(\mu)}$ share the same sign pattern \cite{betteti2024capacity}. For the OAM, $\Gamma(\bm\phi^*)$ is defined by \eqref{eq.oamphaselock}. 

\begin{defin} \label{def.errorfreecap}
Error-free capacity is the largest number of patterns, $K\leq K^{\rm max}_{\rm no \, errors}$, such that
\begin{equation}
    \lim_{N\rightarrow\infty}\mathbb P\left[\bigcap_{\mu = 1}^K \mathcal A(\mu)\right] = 1,
\label{eq.capacityprob}
\end{equation}
where $\cap_{\mu=1}^K \mathcal A(\mu)$ denotes the event in which all patterns are perfectly stored.
\end{defin}

\begin{thm} \label{thm.capacity}
    Let $W = \frac{1}{N} \Xi\Xi^\transp$, where $\bm\xi^{(\mu)}$, $\mu = 1,\ldots,K$, are i.i.d. random variables sampled according to Eq.~\eqref{eq.randsamp}.
    \begin{enumerate}
        \item For the Hopfield model \textnormal{\cite{mceliece2003capacity,betteti2024capacity}}:
        %
            $K^{\rm max}_\textnormal{no errors} \leq \frac{N}{4\log N}$.
        
        \item For the OAM \textnormal{\cite{nishikawa2004capacity}}:
        %
            $K^{\rm max}_\textnormal{no errors} = \frac{2N\kappa^2}{\log N}$.

    \end{enumerate}
\end{thm}

Fig.~\ref{fig.oam}b compares the capacity scaling of the associative memory models considered in this tutorial. In the small-error case (Fig.~\ref{fig.oam}b, left panel), the Hopfield model exhibits linear scaling: $K^{\rm max} \approx 0.14 N$ \cite{amit1985storing}. This capacity is derived for binary randomly-sampled memory patterns, which can increase substantially under additional structural assumptions on the patterns. For instance, if patterns are orthogonal (i.e.,  $\bm\xi^{(\mu)\transp} \bm\xi^{(\mu')} = 0$, $\forall\mu\neq\mu'$), then $K^{\rm max} = N$ \cite{horn1987hypercubic}. If patterns are sparse vectors (i.e., only a fraction $f_{\rm s}\ll 1$ of neurons are active in each pattern), then $K^{\rm max} \sim N/({f_{\rm s} |\log f_{\rm s}|})$ \cite{tsodyks1988enhanced}. The latter result is particularly relevant in practice since biological neural representations are empirically observed to be sparse \cite{BAO-DJF:96}. Because the variance of crosstalk noise between sparse vectors is smaller, the capacity is effectively larger for sparser patterns compared to dense ones\textemdash although the scaling with $N$ remains linear. Superlinear scaling is achieved for DenseAMs with higher-order energy functions $F$ of degree $n\geq 3$, and can even become exponential when $F$ is chosen to be exponential.

For $\kappa\gg 0$, the OAM dynamics effectively reduce to a binary-state Hopfield system, yielding a small-error capacity of the same order. In contrast, when $\kappa=0$, the OAM reduces to a Kuramoto network, which is known to have significantly lower scaling: $K^{\rm max} \approx 0.04 N$ \cite{cook1989mean,aonishi1998phase,yamana1999oscillator}. Recent work has shown that, for a specific network topology known as 1D honeycombs, the Kuramoto network can exhibit an exponentially large number of equilibrium points as a function of the network size $N$ \cite{guo2025oscillatory}. This result can potentially overcome the limited capacity of phase-oscillator models, but it requires the prior specification of a bijective mapping $\Gamma$ between equilibria $\bm x^*$ and patterns $\bm\xi^{(\mu)}$.

In the error-free case (Fig.~\ref{fig.oam}, right panel), the capacity of both Hopfield and OAM models grows sublinearly as a function of $N$, as established in Theorem~\ref{thm.capacity}. Thus, the relative capacity per neuron, $K^{\rm max}_\text{no errors}/N$, tends to zero as $N\rightarrow\infty$. In the Hopfield network, $K^{\rm max}_\text{no errors}$ represents an upper bound since the capacity can be lower depending on the choice of activation function \cite{betteti2024capacity} or the lower bound on the size of the basins of attraction \cite{mceliece2003capacity}. 
In OAMs, the capacity increases monotonically with the second-harmonic strength $\kappa$. In fact, the next theorem shows that $\kappa$ serves as a tunable parameter that directly determines the equilibrium stability:
\begin{thm}[\cite{nishikawa2004oscillatory}] \label{thm.oamstability}
    Consider the OAM \eqref{eq.oam} with Hebbian weights $W=\frac{1}{N}\Xi\Xi^\transp$. The equilibrium point $\bm\phi^*$ associated with a binary pattern $\bm\xi^{(\mu)}$, via relation~\eqref{eq.oamphaselock}, is asymptotically stable if $\lambda_{\rm max}(J(\bm\xi^{(\mu)})) < 2\kappa$, where 
    \begin{equation}
        J(\bm\xi^{(\mu)}) = \frac{1}{N} \left(D_{\mu} W D_{\mu} - \operatorname{diag}(W\bm 1_N)\right),
    \end{equation}
    $D_{\mu} = \operatorname{diag}(\xi^{(\mu)}_1, \ldots, \xi^{(\mu)}_N)$, $\bm 1_N$ is the $N$-dimensional ones vector, and $\lambda_{\rm max}(\cdot)$ is the largest eigenvalue of a matrix.
\end{thm}


\noindent
Thus, increasing $\kappa$ both enlarges the error-free capacity (per Theorem~\ref{thm.capacity}) and increases the stability margins of stored memory patterns (per Theorem~\ref{thm.oamstability}). 
OAMs can then be tuned with an appropriate choice of $\kappa$ so that only the desired patterns are stable while all other possible binary patterns remain unstable (Fig.~\ref{fig.oam}c, adapted from \cite{nishikawa2004capacity}). In this sense, the second-harmonic signal acts as a tunable stability regulator that eliminates the existence of spurious memories, providing additional control over the energy landscape and the computational properties of oscillator-based systems.

\subsection{Unconventional Computing}
\label{sec.oim}

Several combinatorial optimization problems, including graph partitioning, scheduling, and satisfiability, are of large interest to technological, scientific, and logistic applications \cite{mohseni2022ising}. These problems are often computationally intractable, with complexity that grows exponentially with problem size. A broad class of such problems can be defined on a graph and reformulated in terms of the \textit{Ising Hamiltonian}:
\begin{equation}
    \label{eq.ising}
    \mathcal H(\bm\sigma) = -\frac{1}{2}\sum_{i=1}^{N}\sum_{j=1}^{N}W_{ij}\sigma_i \sigma_j,
\end{equation}
\noindent
where $\bm \sigma\in\{\pm 1\}^N$ denotes a spin configuration and $W\in\{0,-1\}^{N\times N}$ is a binary adjacency matrix associated with the undirected graph $\mathcal G$. Here, note that $W$ is specified by the optimization problem rather than learned. The objective is to find the configuration that minimizes the energy, $\min_{\bm \sigma} \mathcal H(\bm \sigma)$, known as the Ising optimization problem.

\begin{rem}
    The Ising optimization directly maps to the MaxCut problem, which seeks a partition of the nodes into two disjoint sets that maximizes the number of edges between them. Two nodes are assigned to different partitions if their spins differ. Thus, adjacent nodes with opposite spins (so that $W_{ij} x_i x_j=1$) lower the energy $\mathcal H$, favoring partitions with more crossing edges. Since MaxCut is NP-complete, and any NP problem can be reduced to it in polynomial time, the Ising optimization problem provides a unifying representation for a broad class of combinatorial problems \cite{lucas2014ising}.
\end{rem}

As in OAMs, we can map the Ising Hamiltonian \eqref{eq.ising} onto the phase dynamics of coupled oscillators through the energy
\begin{equation}
    \label{eq.phaseenergy}
    \mathcal E(\bm \phi) = -\frac{1}{2}\sum_{i,j=1}^N W_{ij}\cos(\Phi_{ij}) + \kappa \sum_{i=1}^{N} \sin(\phi_i)^2, 
\end{equation}
\noindent 
where $\bm\phi\in\mathbb S^N$ and $\kappa\geq 0$ is a regularization parameter that penalizes states with $\phi_i\notin\{0,\pi\}$. The equilibrium $\bm\phi^*$ encodes a spin configuration $\bm \sigma$ when
\begin{equation}
    \label{eq.splay}
    \phi_i^*(\sigma_i) = 
    \begin{cases}
        0,   \,\,\, &\text{if} \,\,\, \sigma_i = +1, \\
        \pi, \,\,\, &\text{if} \,\,\, \sigma_i = -1,
    \end{cases}
\end{equation}
under which $\mathcal E(\bm\phi^*(\bm \sigma)) = \mathcal H(\bm \sigma)$. 
The EDM $\dot{\bm\phi} = -\nabla \mathcal E(\bm\phi)$ thus leads to the phase-oscillator network:
\begin{equation}
    \label{eq.oim}
    \dot{\phi}_i = \omega + \sum_{j=1}^{N}W_{ij}\sin(\phi_j - \phi_i) - \kappa \sin(2\phi_i).
\end{equation}
\noindent
This system is known as an oscillator Ising machine (OIM) \cite{wang2019oim}: an implicit solver whose states asymptotically converge to candidate solutions of the Ising optimization problem. Because there is now a single second-harmonic signal injection per oscillator (rather than $N$ signals per oscillator as in OAMs), OIMs can be implemented through analog electronic oscillators at scale \cite{mallick2020using}. 

Signed graphs provide a useful representation for the stability analysis of OIMs (see \cite{pan2016laplacian,shi2019dynamics} for general properties of signed graphs). For a given graph $\mathcal G$ and spin configuration $\bm\sigma$, we define a \textit{signed graph} $\mathcal G_{\rm s}$ and its corresponding signed adjacency matrix $A_{ij}(\bm\sigma) = W_{ij}\sigma_i\sigma_j$. Thus, edges in $\mathcal G_{\rm s}$ are negative if adjacent spins are aligned, and positive otherwise. The signed Laplacian matrix is $L(\bm\sigma)=D(\bm\sigma)-A(\bm\sigma)$, where $D(\bm\sigma) = \operatorname{diag}(d_{{\rm in},1}, \ldots, d_{{\rm in},N})$ with node in-degree $d_{{\rm in},i}=\sum_j A_{ij}$. Let $\nabla^2 \mathcal E(\bm\phi^*(\bm\sigma))$ be the Hessian matrix associated with the equilibrium point $\bm\phi^*(\bm\sigma)$. This equilibrium is stable if $\lambda_{\rm min}(\nabla^2 \mathcal E)>0$, where $\lambda_{\rm min}(\cdot)$ denotes the smallest eigenvalue of a matrix. The key intuition is that, for each spin configuration $\bm\sigma$, both the Ising Hamiltonian and the OIM stability are determined by the spectral properties of the signed Laplacian \cite{allibhoy2025global}:
\begin{equation}
\begin{aligned}
    \mathcal H(\bm\sigma) &= - \frac{1}{2}\operatorname{tr}(L(\bm\sigma)) &\quad \text{(optimality)}, \\
    \nabla^2E(\bm\sigma) &= L(\bm\sigma) + 2\kappa &\quad \text{(stability)}.   
\end{aligned}
\end{equation}

\noindent
The first relation determines the optimality of a spin configuration, while the second determines its stability. Once again, the second-harmonic strength $\kappa$ serves as a control parameter that determines which equilibria are stable and, therefore, accessible to the OIM relaxation dynamics. 

The following theorem characterizes the relationship between optimality and stability in a statistical sense for random matrices. First, let us define $\mathcal G(N,p)$ as the ensemble of $N$-node random signed graphs $\mathcal G_{\rm s}$ such that the network topology is determined by an Erdős-Renyi graph (i.e., $W_{ij}=-1$ with probability $p$ and 0 otherwise) and the spin configuration is uniformly sampled from $\{\pm 1\}$. Thus,
\begin{equation}
    A_{ij}(\bm\sigma) =
    \begin{cases}
        0, \quad &\text{with probability } (1-p), \\
        -1, \quad &\text{with probability } p/2,  \\
        +1, \quad &\text{with probability } p/2.  \\
    \end{cases}
\end{equation}

\begin{thm}[\cite{allibhoy2025global}]
    Given the ensemble of random graphs $\mathcal G(N,p)$, the conditional expected value of the eigenvalues $\lambda$ of $\nabla^2 \mathcal E(\bm\phi^*(\bm\sigma))$ given an energy level $\mathcal H(\bm\sigma) = h$ is
    \begin{equation}
        \mathbb E[\lambda | \mathcal H(\bm\sigma) = h] = - \frac{2}{N}h + 2\kappa.
    \end{equation}
\end{thm}

\noindent
Therefore, low-energy states tend to be more stable than high-energy ones. As a result, the parameter $\kappa$ serves as a knob that can stabilize very low-energy states while it destabilizes higher-energy ones \cite{bashar2023stability}. In this regime, the convergence of OIMs becomes biased toward (near-)global minima, enhancing their effectiveness as implicit optimization solvers. Prior work has further improved convergence by introducing oscillator heterogeneity on $\kappa$ \cite{allibhoy2025global}.

Oscillator networks have found increasing attention in unconventional computing and combinatorial optimization because their dynamics map naturally onto electronic and photonic hardware implementations \cite{mallick2020using,al2025programmable,khan2025analyzing}. For both OAMs and OIMs, previous studies have characterized how the basins of attraction depend on the second-harmonic signal \cite{cheng2024control,wang2025distributed}, noise \cite{du2024active}, and higher-order network interactions \cite{fu2025role,sun2025general}. Recent work has also shown that oscillator-based networks can improve classification performance \cite{keller2023neural,effenberger2025functional}, admit local learning rules \cite{wang2024training,stern2025physical}, and support generative models such as RBMs \cite{ekanayake2026configuring,guzman2025unsupervised,hasan2026top}. Moreover, given that periodically firing neurons can often be dimensionally reduced to phase-oscillator dynamics \cite{brown2004phase,nakao2016phase}, these oscillator models offer a natural abstraction for the analysis and design of spiking neural networks (SNNs), with applications in associative memory \cite{izhikevich2006polychronization,makovkin2025toward} and neuromorphic control \cite{sepulchre2022spiking,schmetterling2024neuromorphic,belaustegui2025tunable}.


\section{Proximal-Gradient Neural Networks for Sparse and Constrained Reconstruction}
\label{sec.proximal}

\subsection{Neural Circuits for Optimization}
\label{subsection:NeuralCircuitsforOptimization}

\newcommand{\Efrancesco}{\mathcal{E}}


A central tool in the analysis developed here is the following \emph{regularized optimization problem},
\begin{equation}
  \min_{\bm x \in \mathbb{R}^N} \quad \Efrancesco_{\mathrm{regularized}}(\bm x, \bm u)
  ~=~ f(\bm x, \bm u) ~+~ g(\bm x),
  \label{eq:reg-opt}
\end{equation}
where the \emph{nominal cost} $f(\bm x, \bm u)$ is assumed to be well-behaved (differentiable and convex in
$\bm x$), while the \emph{regularizer} $g(\bm x)$ may be nonsmooth, unbounded, or otherwise poorly
behaved. The key object associated with such problems is the \emph{proximal operator} of $g$,
defined as
\begin{equation}
  \mathrm{prox}_{g}(\bm x) ~:=~ \argmin_{\bm z \in \mathbb{R}^N} \quad g(\bm z) + \tfrac{1}{2}\|\bm x - \bm z\|_2^2.
  \label{eq:prox}
\end{equation}
Intuitively, $\mathrm{prox}_{g}$ solves a simple regularized subproblem in
which the quadratic penalty keeps the minimizer close to the input $\bm x$; it
thus maps $\bm x$ to a ``$g$-better'' point while not straying too far. The
proximal operator generalizes the notion of projection and is well-defined
for every closed convex proper function $g$ (e.g.,
see \cite{AB:17,NP-SB:14}).

The proximal operator provides the natural building block for
minimizing~\eqref{eq:reg-opt} via continuous-time dynamics. The
\emph{proximal gradient descent} flow is
\begin{equation}
  \dot{\bm x} ~=~ -\bm x ~+~ \mathrm{prox}_{g}\bigl(\bm x - \nabla_{\bm x} f(\bm x, \bm u)\bigr)
  ~=:~ \mathsf{F}_{\mathrm{ProxG}}(\bm x, \bm u).
  \label{eq:proxG}
\end{equation}
This is an energy system, that is, an EDM, fully determined by the pair
$(f, g)$, in direct analogy with the ordinary gradient descent $\dot{\bm x} =
-\nabla_{\bm x} f$ in \eqref{eq.generaledm}, which is determined by $f$
alone. This approach was proposed and formalized in \cite{AG-AD-FB:24d},
building on the continuous-time proximal flow introduced
in \cite{SHM-MRJ:21}.

\begin{result}[\textit{Proximal gradient descent equals the firing-rate network}]
Comparing system~\eqref{eq:proxG} with the FRN~\eqref{eq.ratedyn}, one observes that the two are identical
whenever $f$ is quadratic in $(\bm x, \bm u)$ and the activation function satisfies $\Phi(\bm x) =
\mathrm{prox}_{g}(\bm x)$. That is, the FRN is precisely the proximal gradient descent
for a regularized energy whose structure is encoded in $\Phi$ and $W$.
\end{result}

\begin{result}[\textit{The Hopfield energy is a regularized energy}]
More precisely, the FRN is the proximal gradient
descent for the \emph{Hopfield energy}~\eqref{eq.hopfield_energy}, which (after an
appropriate rewriting~\cite{SB-WR-AD-JC-FB:25i,SB-GB-FB-SZ:24m}) admits the regularized decomposition
\begin{equation}\label{eq.regularizeddecomp}
  \Efrancesco_{\mathrm{regularized}}(\bm x, \bm u)
  ~=~ \Efrancesco_{\mathrm{network}}(\bm x, \bm u)
  ~+~ \sum_{i=1}^{N} \Efrancesco_{\mathrm{activation},i}(x_i).
\end{equation}
The two components play distinct roles. The \emph{network energy}
\begin{equation}\label{eq.networkenergy}
  \Efrancesco_{\mathrm{network}}(\bm x, \bm u)
  ~=~ \tfrac{1}{2} \bm x^\top (I_n - W) \bm x - \bm x^\top B \bm u
\end{equation}
captures pairwise synaptic interactions and the effect of the external stimulus. The
\emph{activation energy} $\Efrancesco_{\mathrm{activation},i}$ determines the activation function of
each neuron $i$ via $\Phi_i(y) = \mathrm{prox}_{\Efrancesco_{\mathrm{activation},i}}(y)$, so that
each choice of regularizer corresponds to a distinct biophysical nonlinearity.

As a concrete example, the indicator function $g(x) = \iota_{[0,1]}(x)$ (equal to $0$ if
$0 \leq x \leq 1$ and $+\infty$ otherwise) has proximal operator equal to the saturation
function $\mathrm{sat}_{0,1}(x) = [x]_0^1$. The resulting network is the \emph{linear threshold
model},
\begin{equation}\label{eq.linearthreshold}
  \dot{\bm x} = -\bm x + [W\bm x + B\bm u]_0^1.
\end{equation}
This is a widely used model in computational neuroscience whose activation nonlinearity thus emerges from a
simple box-constraint regularizer on the state space (e.g., see the early ref. \cite{KPH:74}).
\end{result}

\begin{result}[\textit{Dynamical systems analysis}]
The vector field $\mathsf{F}_{\mathrm{ProxG}}$ enjoys several analytically
favorable properties, many of which established
in~\cite{AG-AD-FB:24d}. First, it is well-posed and Lipschitz, and is
uniquely determined by the pair $(f, g)$. Second, the optimization and
dynamical perspectives are equivalent: a point $\bm x^*$ minimizes $f + g$ if
and only if it is an equilibrium of $\mathsf{F}_{\mathrm{ProxG}}$. Third,
the regularized cost $f + g$ is non-increasing along every bounded
trajectory of the flow (decreasing energy). Fourth, under the spectral
condition $W \prec I_n$, the flow is contracting \cite{FB:26-CTDS},
which implies global exponential convergence to the unique equilibrium and
various robustness properties. Fifth, convergence rates can be certified
via a proximal Polyak--{\L}ojasiewicz condition, yielding quantitative
bounds on the optimization speed.
\end{result}

\begin{result}[\textit{Analog circuit implementation}]
The proximal gradient dynamics~\eqref{eq:proxG} admit a direct analog hardware realization.  Wu et
al.~\cite{JW-XH-YN-HT-JZ:24} show that the flow can be implemented with three operational amplifiers
per state dimension, provided the gradient $\nabla f$ and the proximal operator are implementable.
This result establishes a concrete bridge between our mathematical framework and neuromorphic
circuit design.
\end{result}

Taken together, these results support a unified normative framework: firing-rate dynamics can be
re-interpreted as proximal gradient dynamics defined by a regularized energy, with the network
energy describing synaptic interactions and the activation energy encoding physical constraints on
firing rates. This constitutes an optimization-based, top-down framework that \emph{derives} neural
circuit equations from a mathematical objective, and naturally requires symmetric synapses ($W =
W^\top$) as a structural consequence of the energy formulation.  We next provide two case studies on
the proximal gradient descent dynamics.

\smallskip
\subsubsection{Case Study 1: Sparse Signal Reconstruction}
A prominent functional hypothesis in systems neuroscience is that the
primary visual cortex (V1) encodes sensory stimuli in a sparse format. At
any given moment, only a small fraction of neurons are active, and the
receptive fields that serve as the encoding dictionary are learned from the
statistical structure of natural scenes, see the landmark studies~\cite{BAO-DJF:96, BAO-DJF:04}.  The same architectural principle appears
in invertebrates: in the mushroom body of the locust, Kenyon cells project
excitatorily onto a single GABAergic interneuron, which in turn inhibits
all Kenyon cells globally, implementing a normalization mechanism that
enforces sparse odor representations~\cite{MP-SC-TN-GL:11}.  Sparsity also
plays a central role in signal processing and machine learning, where it
underlies compressed sensing, image denoising, and dimensionality
reduction~\cite{EJC-TT:05, JW-YM:22}.

The mathematical formulation of this problem is based on the \emph{positive lasso} cost function:
\begin{equation}
  \min_{\bm x \in \mathbb{R}^N,\; \bm x \geq 0} \quad \Efrancesco_{\mathrm{lasso}}(\bm x)
  ~:=~ \|\bm u - \Theta \bm x\|_2^2 ~+~ \lambda \|\bm x\|_1,
  \label{eq:lasso}
\end{equation}
where $\bm x$ is constrained to be $k$-sparse and nonnegative, $\Theta = [\Theta_1 \, \ldots \, \Theta_n] \in \mathbb{R}^{M \times N}$ is an overcomplete dictionary ($k \ll M \ll N$) with
unit-norm columns $\|\Theta_i\| = 1$, and the inner products $\Theta_i \cdot \Theta_j$ measure similarity
between dictionary atoms.  The quadratic
term penalizes reconstruction error, while the $\ell_1$ regularizing term is well known to promote
sparsity.

\begin{figure}[t]\centering
  \includegraphics[width=\linewidth]{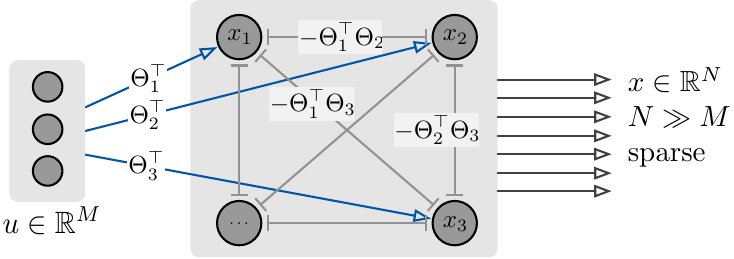}
  \caption{Positive competitive network implementing sparse signal
    reconstruction via the proximal gradient
    dynamics. Each excitatory neuron $i$ receives
    feedforward drive $\Theta_i^\top \bm u$ proportional to its dictionary match,
    and is laterally inhibited by neuron $j$ with strength ${-}\Theta_i^\top
    \Theta_j\leq0$ (for nonnegative dictionaries). A global bias $\lambda$
    controls the sparsity level of the steady-state
    representation.}\label{fig:nn-competitive}
\end{figure}

Applying proximal gradient dynamics to~\eqref{eq:lasso} yields a
\emph{positive competitive network} \cite{VC-AG-AD-GR-FB:23a}:
\begin{equation}
  \dot{\bm x} ~=~ -\bm x ~+~ \operatorname{ReLU}\Bigl((I_N - \Theta^\top\Theta)\,\bm x + \Theta^\top \bm u - \lambda
  \mathbf{1}_N\Bigr).
  \label{eq:sparse-net}
\end{equation}
This dynamics, proposed in \cite{VC-AG-AD-GR-FB:23a},
is inspired by and consistent with the early works~\cite{CJR-DHJ-RGB-BAO:08, AB-JR-CJR:12}. The
biological interpretation of~\eqref{eq:sparse-net} becomes transparent when the dynamics are written
neuron-by-neuron, as illustrated in Fig.~\ref{fig:nn-competitive}.  For each neuron $i$,
\begin{equation*}
  \dot{x}_i ~=~ -x_i ~+~ \mathrm{ReLU}\Bigl(
    \sum_{j \neq i} \underbrace{(-\Theta_i^\top \Theta_j)}_{\leq 0,\;\text{lateral inhibition}} \hspace{-10pt} x_j
    ~+~ \underbrace{\Theta_i^\top \bm  u}_{\text{stimulus}}
    ~-~ \underbrace{\lambda}_{\text{bias}}
  \Bigr).
\end{equation*}

\noindent
Each neuron receives feedforward drive proportional to its match with the stimulus~$\bm u$, and is
suppressed by other neurons in proportion to their dictionary overlap\textemdash a form of direct lateral
inhibition. The network thus performs sparse signal reconstruction through biologically
plausible local interactions, with the sparsity level controlled by the bias~$\lambda$.

\smallskip
\subsubsection{Case Study 2: Policy Composition via Free Energy} We now briefly review a second application of the proximal gradient framework proposed
in \cite{FR-VC-FB-GR:25k}, that connects to the \emph{free energy principle} in
neuroscience~\cite{KF:10,AS-HJ-KF-GR:25}. The starting point is probabilistic mind theory, which
posits that the brain represents information as probability distributions and uses Bayesian
inference for perception, learning, and decision-making. Within this framework, adaptive behavior in
both natural and artificial agents arises from the minimization of a \emph{variational free energy}
(which we here approximately equate with a notion of ``surprise''): perception corresponds to
adjusting beliefs via variational Bayesian inference, learning corresponds to updating generative
models, and decision-making corresponds to acting so as to bring sensory inputs into alignment with
predictions.

The \emph{policy composition problem} provides a concrete instantiation of this principle. Given a
set of primitive policies $\{\text{primitive}_\alpha(\bm u \mid \bm x)\}$, one seeks to obtain a mixture policy
\begin{equation}
  \text{policy}(\bm u \mid \bm x) ~=~ \sum_\alpha w_\alpha \; \text{primitive}_\alpha(\bm u \mid \bm x)
\end{equation}
by minimizing the free energy objective
\begin{equation}
  \min_{\substack{\bm w \geq 0, \\ \mathbf{1}^\transp \bm w = 1}} \quad
  \underbrace{\text{surprise}(\bm x, \bm u)}_{\text{prior belief vs.\ actual outcomes}}
  \hspace{-15pt}~-~ \tau \, \underbrace{\text{entropy}(\bm w)}_{\text{uncertainty}},
  \label{eq:free-energy}
\end{equation}
where $\tau > 0$ is a temperature parameter trading off surprise minimization against uncertainty
maximization. Applying proximal gradient dynamics to \eqref{eq:free-energy} yields a \emph{softmax
gradient descent} FRN~\cite{FR-VC-FB-GR:25k}:
\begin{equation}
  \dot{\bm w} ~=~ -\bm w ~+~ \mathrm{softmax} \bigl(-\frac{1}{\tau} \nabla \,\text{surprise}(\bm x, \bm w)\bigr).
  \label{eq:softmax-net}
\end{equation}
The entropy regularizer in \eqref{eq:free-energy} plays the role of the function $g$ in the general
proximal framework, with the softmax function emerging as its proximal operator on the probability
simplex (see \cite{FR-VC-FB-GR:25k} for a careful derivation). This demonstrates that the normative
framework extends naturally beyond quadratic energies to information-theoretic objectives, yielding
biologically interpretable dynamics from the top-down derivation principle.

\subsection{Neural Circuits for Multiplayer Optimization}

The analysis in Sec.~\ref{subsection:NeuralCircuitsforOptimization} relies on the symmetry of
the synaptic weight matrix $W$, which enables the interpretation of the FRN as
gradient or proximal gradient descent for a single scalar energy. Biological neural circuits,
however, routinely violate this symmetry, as discussed in Sec.~\ref{sec.hopfield}. According to Dale's law, each neuron exerts the same type of effect\textemdash excitatory or inhibitory\textemdash on all of its postsynaptic targets. (In a neuromodulatory formulation, each neuron releases the
same neurotransmitter at all of its synapses.) As a consequence, biologically realistic networks are
naturally described by separate populations of excitatory (E) and inhibitory (I) neurons with
asymmetric connectivity, including classic circuit motifs such as the Wilson--Cowan
excitatory-inhibitory pair~\cite{wilson1972excitatory} and the $\mathrm{E}^k$-I architecture in which a central inhibitory neuron mediates competition among $k$ excitatory neurons. The populations of excitatory and inhibitory neurons are respectively denoted by the sets ${\rm E}\subset \{1,\ldots,N\}$ and ${\rm I} = \{1,\ldots,N\}\backslash{\rm E}$.

For such asymmetric \emph{E-I networks}, the full range of nonlinear dynamical behaviors becomes
possible: global asymptotic stability, multistability, limit cycles, and chaos. Yet despite their
biological ubiquity, these networks have lacked a general analysis framework for stability and
functionality, and a systematic design framework analogous to the optimization-based approach of
Sec.~\ref{subsection:NeuralCircuitsforOptimization}. The results of this section, proposed
in \cite{SB-WR-AD-JC-FB:25i}, extend the proximal gradient framework to a multiplayer setting and
provide a first step towards addressing both gaps.

\begin{result}[\textit{Neural circuits as multiplayer optimization and proximal gradient play}]
The key conceptual shift is to abandon the single-energy perspective and instead assign to
each neuron $i$ an individual cost function. Recall that, for symmetric networks, the FRN dynamics \eqref{eq.ratedyn} is proximal gradient descent with respect to the regularized energy \eqref{eq.regularizeddecomp}.
For asymmetric networks,
the same dynamical equation is reinterpreted as \emph{proximal gradient play}: each neuron $i$
simultaneously minimizes its own individual cost
\begin{multline*}
  \Efrancesco_{\mathrm{regularized},i}(x_i, x_{-i}, \bm u)
  = \Efrancesco_{\mathrm{individual},i}(x_i, x_{-i}, \bm u) \\
  + \Efrancesco_{\mathrm{activation},i}(x_i),
\end{multline*}
where
\begin{equation*}
  \Efrancesco_{\mathrm{individual},i}(x_i, x_{-i}, \bm u)
  ~=~ \sum_{j=1}^N \Bigl(\tfrac{1}{2}\delta_{ij} - 1\Bigr) W_{ij}\, x_i x_j - \bm x^\top B\bm u,
\end{equation*}
and $x_{-i}$ denotes the states of all neurons other than $i$. Under this interpretation,
equilibrium points of the FRN are precisely the Nash equilibria of the game
defined by the individual costs $\{\Efrancesco_{\mathrm{regularized},i}\}_{i=1}^N$. The transition
from a single shared energy to a collection of individual energies is illustrated geometrically
by the shift from a single landscape with multiple local minima to two distinct landscapes
whose intersection defines the Nash equilibrium~\cite{SB-WR-AD-JC-FB:25i}.
\end{result}

\begin{result}[\textit{Monostability for E-I networks}]
Consider the linear threshold network \eqref{eq.linearthreshold} satisfying Dale's law, so that
each neuron belongs to either the excitatory class $\mathrm{E}$ or the inhibitory class
$\mathrm{I}$. Assume that E-I connections are \emph{reciprocal} (that is, $(i,j)$ is an edge if and
only if $(j,i)$ is an edge for $i \in \mathrm{E}$, $j \in \mathrm{I}$), and that synaptic weights
are \emph{homogeneous} within each population type: $w_{\mathrm{EE}}$ for every E-to-E synapse,
$w_{\mathrm{EI}}$ for every I-to-E synapse, $w_{\mathrm{IE}}$ for every E-to-I synapse, and
$w_{\mathrm{II}}$ for every I-to-I synapse. Under these structural assumptions, the network is
\emph{monostable}\textemdash meaning a unique equilibrium exists and is globally asymptotically stable\textemdash
when
\begin{equation*}
  \Bigl(\frac{d_{\mathrm{in}} + d_{\mathrm{out}}}{2}\Bigr) w_{\mathrm{EE}} < 1
  \quad\text{and}\quad
  \Bigl(\frac{d_{\mathrm{in}} + d_{\mathrm{out}}}{2} - 2\Bigr) w_{\mathrm{II}} < 1,
\end{equation*}
where $d_{\mathrm{in}}$ and $d_{\mathrm{out}}$ denote the in- and out-degree of the neurons in
the network~\cite{SB-WR-AD-JC-FB:25i}. These conditions bound the excitatory and inhibitory
recurrent gains in terms of the network connectivity, providing a tractable spectral-like
criterion that does not require full knowledge of the weight matrix.
\end{result}

\begin{result}[\textit{Functionality via lateral inhibition}]
Under the same Dale's law structure, the network also exhibits rich and interpretable functional
behavior when the synaptic weights additionally satisfy the \emph{functionality condition}
$w_{\mathrm{IE}} \geq 1 + w_{\mathrm{II}}$, which controls the strength of feedforward inhibition
relative to inhibitory self-coupling.

In the minimal $\mathrm{E}^2$-I motif (two excitatory neurons, left and right, interconnected to a 
common inhibitory interneuron), lateral inhibition produces binary decisions. Specifically, defining the
threshold $\delta := 1 - w_{\mathrm{EE}} + w_{\mathrm{EI}} > 0$, one has: if
the stimulus to the left neuron satisfies $u_{\mathrm{left}} > \delta$ and 
the stimulus to the right neuron satisfies $u_{\mathrm{right}} < -\delta$, then the left E neuron
settles to a high firing rate and the right E neuron to a low one; and vice-versa when the
stimulus inequality is reversed.
This lateral-inhibition mechanism generalizes to the $\mathrm{E}^k$-I architecture with $k$
excitatory neurons competing through a shared inhibitory pool (Fig.~\ref{fig:nn-WilsonCowan-EkI}), where it implements a \emph{winner-take-all} computation: if the vector of stimuli satisfies
$u_i > \delta$ and $u_j < -\delta$ for all $j \neq i$, then the $i$th  $\mathrm{E}$ neuron wins 
(high firing rate) and all others are suppressed.

\begin{figure}[t]\centering
  \includegraphics[width=.55\linewidth]{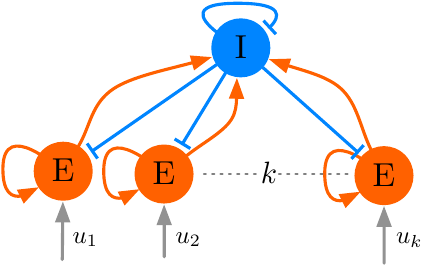}
  \caption{E$^k$-I network implementing winner-take-all competition. A set of $k$
    excitatory neurons project to a shared inhibitory interneuron, which in
    turn inhibits all excitatory neurons. Under the monostability condition
    $w_{\mathrm{EE}} < 1$ and the functionality condition $w_{\mathrm{IE}}
    \geq 1 + w_{\mathrm{II}}$, the network selects the neuron receiving the
    strongest input while suppressing all
    others.}\label{fig:nn-WilsonCowan-EkI}
\end{figure}

Finally, stacking columns of $\mathrm{E}^2$-I motifs into a layered network (with the tighter
monostability condition $w_{\mathrm{EE}} < \tfrac{1}{2}$) enables \emph{contrast enhancement}. Given
an initial input contrast $u_{\mathrm{left}} > u_{\mathrm{right}} + 2\varepsilon$ for
small $\varepsilon > 0$, the contrast is amplified geometrically across layers. After a number
of layers
\begin{equation}
  \ell ~\geq~ \ell_{\mathrm{binary}}
  ~:=~ 1 + \frac{\ln(\varepsilon/\delta)}{\ln(1/w_{\mathrm{EE}} - 1)},
\end{equation}
the output satisfies $u_{\mathrm{left}} > u_{\mathrm{right}} + \delta$,
achieving full binary contrast enhancement~\cite{SB-WR-AD-JC-FB:25i}. The
threshold $\ell_{\mathrm{binary}}$ is determined entirely by the synaptic
weights $w_{\mathrm{EE}}$ and the initial contrast $\varepsilon$, providing
a quantitative design criterion for contrast-enhancing neural
architectures.
\end{result}

Together, these results establish a unified framework for stability
and functionality in E-I circuits, with contrast enhancement as a canonical
example of emergent computation.

\section{Concluding Remarks}
\label{sec.conclusion}

The EDMs presented in this tutorial highlight a class of computational systems whose dynamics are both inherently amenable to analysis and expressive for a wide range of AI and optimization tasks. Because these models evolve according to well-defined energy functions, we have shown that many control-theoretic tools can be applied for formal guarantees on stability, convergence, and safety. For instance, prior work has developed methods for constraint satisfaction in gradient-flow systems \cite{allibhoy2023control,gadginmath2026provably}, improved convergence rates in gradient-based optimization \cite{min2023convergence,zheng2024dissipative}, and optimal data-driven control of adaptive neural systems \cite{moore2024neuron}.
These mathematical developments also provide a foundation for modeling biological learning and inference mechanisms, including the role of stimuli in associative memory \cite{betteti2025input,cornelius2013realistic}, agent heterogeneity in decision-making \cite{gast2024neural,montanari2025optimal,effenberger2025functional,dahmen2025heterogeneity}, and short-term plasticity in working memory \cite{kozachkov2022robust}. At the same time, EDMs typically retain a structure that is naturally compatible with analog hardware implementations \cite{krotov2023new,mallick2020using,JW-XH-YN-HT-JZ:24} as well as physics-informed machine learning \cite{karniadakis2021physics}.
These classic and modern developments position EDMs as promising computing systems that are interpretable, reliable, scalable, and biologically plausible.




\end{document}